\documentclass{article}

\usepackage{arxiv}

\usepackage[utf8]{inputenc} 
\usepackage[T1]{fontenc}    
\usepackage{hyperref}       
\usepackage{url}            
\usepackage{booktabs}       
\usepackage{amsfonts}       
\usepackage{nicefrac}       
\usepackage{microtype}      
\usepackage{lipsum}
\usepackage{graphicx}
\graphicspath{ {./images/} }

\usepackage{graphicx}
\usepackage{amsmath}
\usepackage{amssymb}
\usepackage{booktabs}

\usepackage{capt-of} %

\usepackage[capitalize]{cleveref}
\crefname{section}{Sec.}{Secs.}
\Crefname{section}{Section}{Sections}
\Crefname{table}{Table}{Tables}
\crefname{table}{Tab.}{Tabs.}

\usepackage[titletoc,toc,title]{appendix}
\usepackage{multirow}
\usepackage{microtype}
\usepackage{cases}

\usepackage{amsthm}

\usepackage[linesnumbered,ruled,vlined]{algorithm2e}

\usepackage{booktabs,siunitx}
\newcommand{\mc}[3]{\multicolumn{#1}{#2}{#3}}
\usepackage{adjustbox}

\usepackage{caption}
\usepackage{subcaption}

\usepackage{float}
\usepackage{bm}

\title{Improving Decoupled Posterior Sampling for Inverse Problems using Data Consistency Constraint}

\author{
 Zhi Qi \\
  ZJU-UIUC Institute\\
  Zhejiang University, China\\
  \texttt{zhi1.24@intl.zju.edu.cn} \\
   \And
 Shihong Yuan \\
  ZJU-UIUC Institute\\
  Zhejiang University, China\\
  \texttt{shihong.23@intl.zju.edu.cn} \\
  \And
 Yulin Yuan \\
  ZJU-UIUC Institute\\
  Zhejiang University, China\\
  \texttt{yulinyuan@zju.edu.cn} \\
  \And
 Linling Kuang \\
  School of Information Science and Technology\\
  Tsinghua University, China\\
  \texttt{kll@mail.tsinghua.edu.cn} \\
  \And
 Yoshiyuki Kabashima \\
  Institute for Physics of Intelligence, Department of Physics\\
  The University of Tokyo, Japan\\
  \texttt{kaba@phys.s.u-tokyo.ac.jp} \\
  \And
 Xiangming Meng \\
  ZJU-UIUC Institute\\
  Zhejiang University, China\\
  \texttt{xiangmingmeng@intl.zju.edu.cn} \\
}

\begin{document}

\SetKwInput{KwInput}{Input}                
\SetKwInput{KwInitialize}{Initialization}                
\SetKwInput{KwOutput}{Output}              

\maketitle

\begin{figure*}[!h]

     \centering
     \begin{subfigure}[b]{0.48\textwidth}
         \centering
         \includegraphics[width=\textwidth]{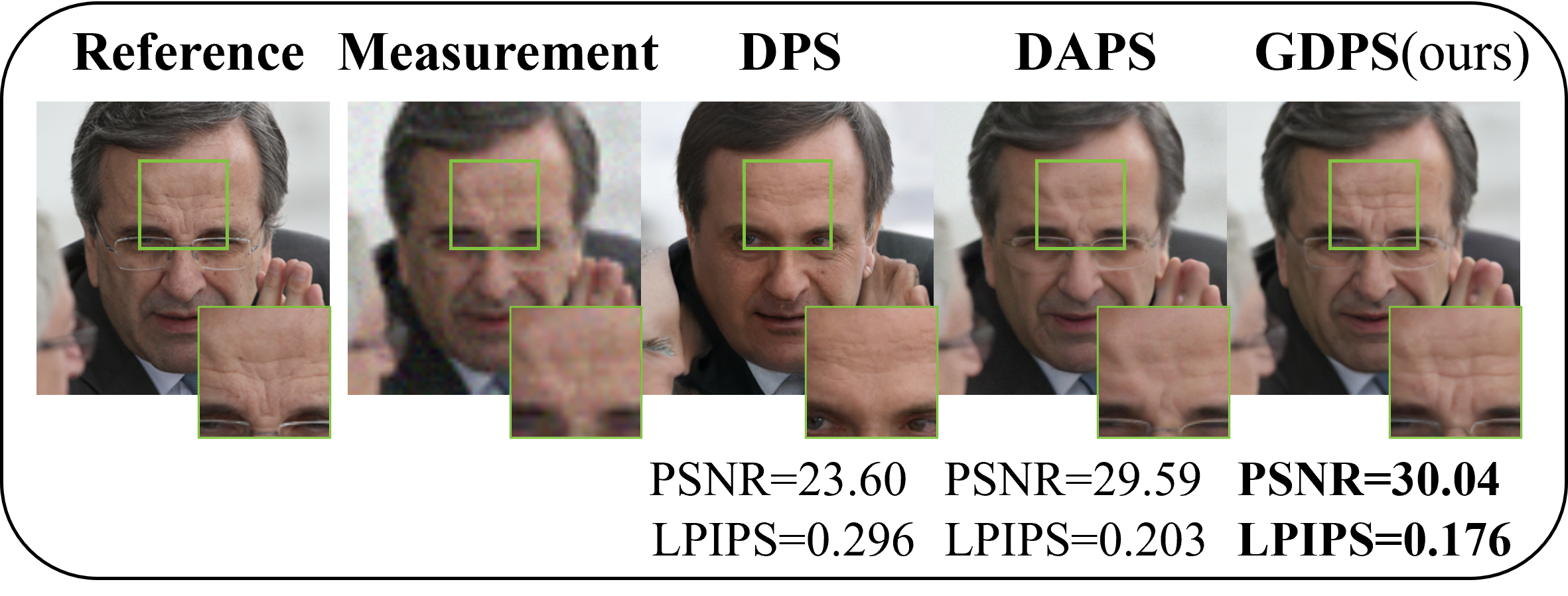}
         \caption{Super Resolution}
     \end{subfigure}
     \hfill
     \begin{subfigure}[b]{0.48\textwidth}
         \centering
         \includegraphics[width=\textwidth]{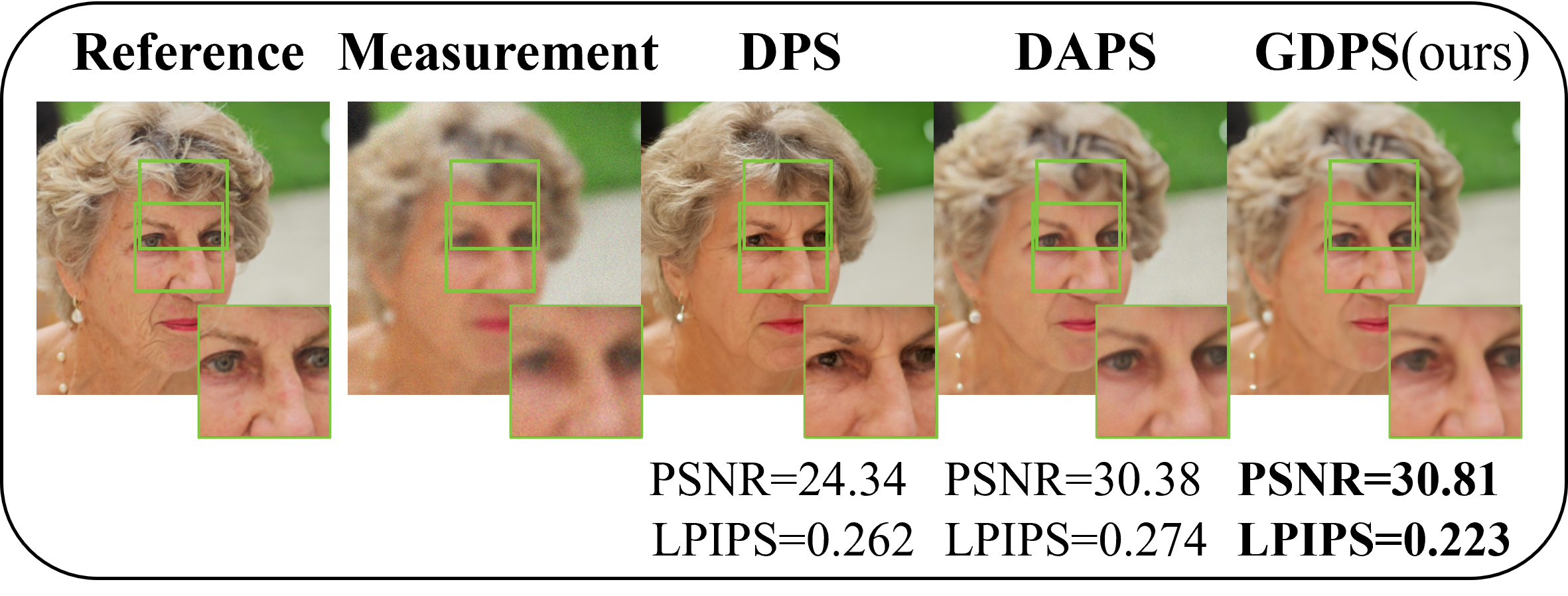}
         \caption{Gaussian Deblurring}
     \end{subfigure}
     \vskip\baselineskip
     \begin{subfigure}[b]{0.48\textwidth}
         \centering
         \includegraphics[width=\textwidth]{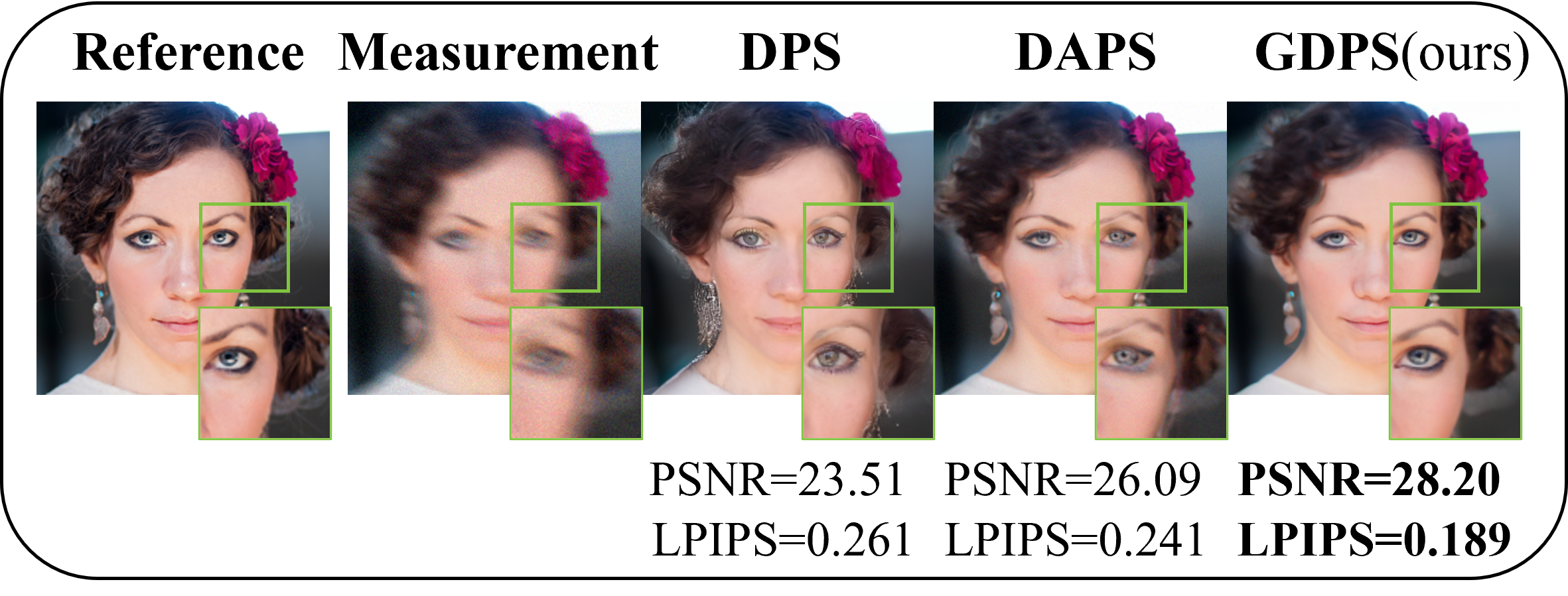}
         \caption{Nonlinear Deblurring}
     \end{subfigure}
     \hfill
     \begin{subfigure}[b]{0.48\textwidth}
         \centering
         \includegraphics[width=\textwidth]{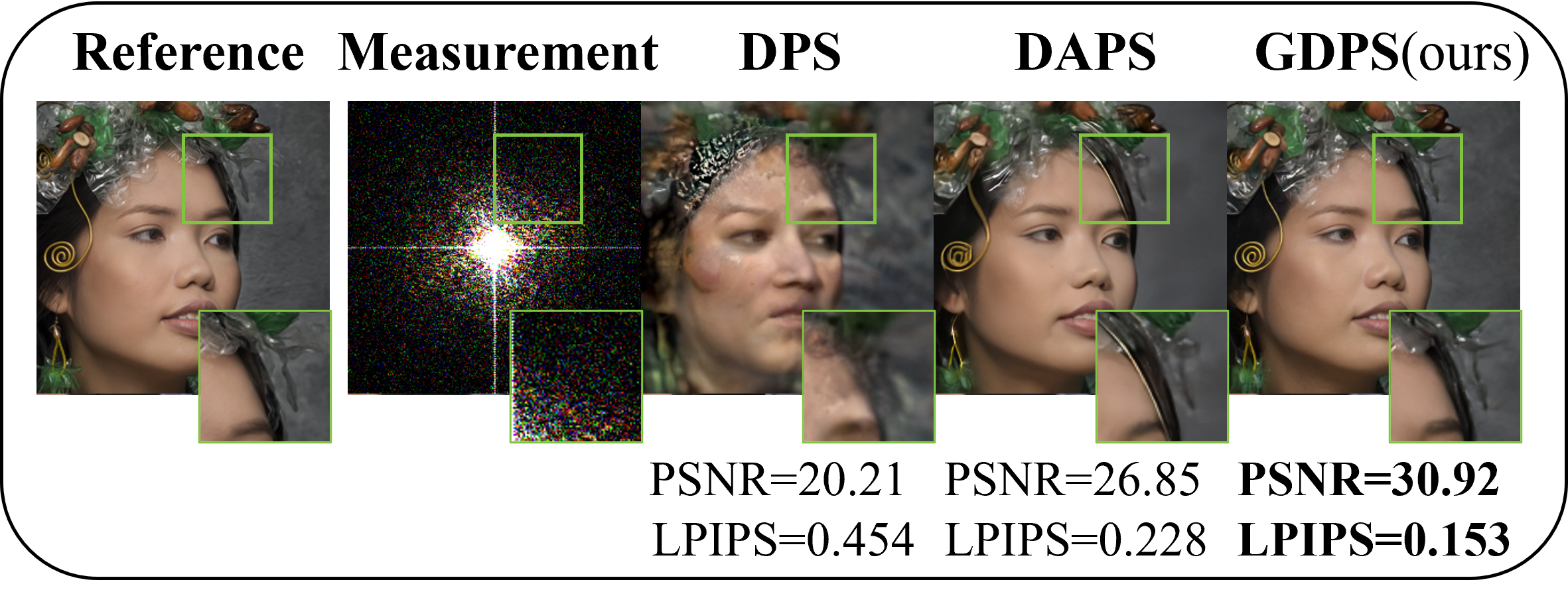}
         \caption{Phase Retrieval}
     \end{subfigure}
    \caption{Representative Results for the DPS, DAPS, and GDPS Methods. The tasks are presented in the following order: (a) super resolution, (b) Gaussian deblurring, (c) nonlinear deblurring, and (d) phase retrieval. As shown in Figure 1, our GDPS method consistently achieves higher quality results compared to other methods. In Figure 1(a), our method successfully reconstructs clear wrinkles from the measurements, enhancing facial texture. In Figure 1(b), GDPS accurately recovers fine details around the eye area, including the eye bags and eye socket, demonstrating improved facial feature restoration. In Figure 1(c), our method effectively restores detailed elements such as the eye shadow from the measurements. Finally, in Figure 1(d), GDPS reconstructs intricate details of the headwear and background with higher clarity. Overall, the visual quality of GDPS’s results is closer to the reference images, underscoring its effectiveness in enhancing clarity and preserving fine details.}
    \label{fig-ffhq}
\end{figure*}

\begin{abstract}
Diffusion models have shown strong performances in solving inverse problems through posterior sampling while they suffer from  errors during earlier steps. To mitigate this issue, several Decoupled Posterior Sampling methods have been recently proposed. However, the reverse process in these methods ignores measurement information, leading to errors that impede effective optimization in subsequent steps. To solve this problem, we propose Guided Decoupled Posterior Sampling (GDPS) by integrating a data consistency constraint in the reverse process. The constraint performs a smoother transition within the optimization process, facilitating a more effective convergence toward the target distribution. Furthermore, we extend our method to latent diffusion models and Tweedie's formula, demonstrating its scalability. We evaluate GDPS on the FFHQ and ImageNet datasets across various linear and nonlinear tasks under both standard and challenging conditions. Experimental results demonstrate that GDPS achieves state-of-the-art performance, improving accuracy over existing methods.
\end{abstract}


\section{Introduction}
Inverse problems are typically defined by the mathematical model $\mathbf{y} = \mathcal{A}(\mathbf{x}) + \mathbf{n}$, where $\mathbf{y}$ represents the noisy measurement, $\mathcal{A}$ is a known operator, $\mathbf{x}$ is the signal to be reconstructed from $\mathbf{y}$, and $\mathbf{n}$ denotes noise. Because inverse problems are inherently many-to-one, they are often ill-posed, lacking a unique solution. Various methods\cite{o1986statistical, candes2006robust, candes2008introduction, fazel2008compressed, foygel2014corrupted, tang2009performance, huang2005inverse} have been proposed to address this issue, many of which rely on signal processing and optimization techniques. However, these approaches often suffer from a lack of prior knowledge, which can cause the reconstructed signals to deviate from the true data distribution.

Diffusion models\cite{ho2020denoising, nichol2021improved, song2019generative, song2020improved, dhariwal2021diffusion, sohl2015deep} address this limitation by introducing a prior distribution into the generation process. Specifically, diffusion models employ a sequence of denoising steps that gradually transform random noise into structured data, making them well-suited for reconstructing signals from noisy measurements. Diffusion models generate samples from a distribution $p(\mathbf{x})$, and through Bayes’ theorem, this prior distribution can be transformed into the posterior distribution $p(\mathbf{x} | \mathbf{y})$ as $p(\mathbf{x} | \mathbf{y}) = \frac{p(\mathbf{x}) p(\mathbf{y} \mid \mathbf{x})}{p(\mathbf{y})}$, which allows for sampling from $p(\mathbf{x} | \mathbf{y})$ to generate signals conditioned on the measurements. Several recent works\cite{meng2022quantized, chung2022diffusion, chung2022improving, jalal2021robust, jalal2021instance, kawar2022denoising, kawar2021snips, song2023pseudoinverse, boys2023tweedie, wang2024dmplug, chung2024direct, alkhouri2024diffusion, murata2023gibbsddrm} perform posterior sampling by estimating likelihoods. However, these methods often introduce substantial bias in the early stages of sampling, which can be challenging to correct in later stages.

Decoupled Posterior Sampling methods\cite{zhang2024improving, li2024decoupled, alkhouri2024sitcom} address this bias issue by leveraging a sequence of noisy samples. Unlike Bayesian-based methods that apply Bayes' theorem directly in the reverse process, Decoupled Posterior Sampling methods decompose the sampling process into three distinct phases: the reverse process, the optimization process, and the forward process. The reverse process transforms a noisy sample into a clearer one, the optimization process adjusts this refined sample to more closely align with the true signal distribution $\mathbf{x}_0$ given measurement $\mathbf{y}$, and the forward process reintroduces noise into the sample to get a noisy sample for the next step. The reverse process can involve iterative solutions of the probability flow ODE\cite{karras2022elucidating, liu2022flow, yang2024consistency} or solution of Tweedie’s formula, while the optimization step may employ techniques such as stochastic gradient descent\cite{robbins1951stochastic}, adaptive moment estimation\cite{kingma2014adam} or Langevin dynamics\cite{welling2011bayesian}. The forward process introduces a noise decreasing with step. Through these three processes, Decoupled Posterior Sampling methods progressively reduces noise, creating a sequence of noisy samples that ultimately converge toward a clean sample while mitigating bias from earlier stages.

However, in Decoupled Posterior Sampling methods, the reverse process often lacks sufficient measurement information, resulting in the optimization process that starts from a biased initial point. To address this, we propose the Guided Decoupled Posterior Sampling (GDPS) method by incorporating a data consistency constraint into the reverse process.
\textbf{Contributions:}
\begin{itemize}\setlength{\itemsep}{0pt}
    \item We improve the decoupled posterior sampling methods by proposing a method that incorporates data consistency constraint, enabling a smooth transition in decoupled posterior sampling.
    
    
    \item We extend our method to the reverse process utilizes latent diffusion models and Tweedie’s formula. This adaptability allows GDPS to deliver improvements across a range of Decoupled Posterior Sampling methods.
    
    \item We test our Guided Decoupled Posterior Sampling(GDPS) on FFHQ\cite{karras2019style} dataset and Imagenet\cite{deng2009imagenet} dataset in various linear tasks and  nonlinear tasks under both standard and challenging conditions. Experiments show that our method consistently achieves state-of-the-art performance on all the tasks considered. 
\end{itemize}

\section{Related Works}
\subsection{Inverse Problems}
In recent years, posterior sampling using diffusion models has become a popular approach for solving inverse problems. Several methods\cite{meng2022diffusion, chung2022diffusion, chung2023prompt, rout2024solving, song2023pseudoinverse, song2023solving, chung2023decomposed, zhu2023denoising, rout2024beyond, dou2024diffusion, wu2024principled, sun2024provable, cardoso2023monte, rozet2024learning} employ posterior sampling by estimating the likelihood, utilizing Bayesian inference to transform the prior distribution into the posterior distribution during the reverse process. In DPS\cite{chung2022diffusion}, the likelihood function \( p(\mathbf{y} | \mathbf{x}) \) is computed under the assumption that the signal distribution \( p(\mathbf{x}_0) \) is a Dirac delta function, with \( \delta = \hat{\mathbf{x}}_0(\mathbf{x}_t, t) \) estimated via Tweedie’s formula\cite{kim2021noise2score, efron2011tweedie}. $\Pi$GDM\cite{song2023pseudoinverse} improves on DPS by replacing the Dirac delta distribution with a Gaussian distribution for \( p(\mathbf{x}_0) \). DMPlug\cite{wang2024dmplug} approaches the reverse process as a function minimization problem, and optimizing the objective \( \ell(\mathbf{y}, \mathcal{A}(\mathbf{x})) + \Omega(\mathbf{x}) \), where \( \ell(\mathbf{y}, \mathcal{A}(\mathbf{x})) \) represents the data-fitting loss and \( \Omega(\mathbf{x}) \) serves as a regularization term.

In the latent space, PSLD\cite{rout2024solving} introduces a gluing objective to address error terms caused by the latent structure, aiming to minimize the discrepancy $\nabla_{\mathbf{z}_t} \left\| \mathbb{E}[\mathbf{z}_0 | \mathbf{z}_t] - \mathcal{E} \left( \mathcal{D} \left( \mathbb{E}[\mathbf{z}_0 | \mathbf{z}_t] \right) \right) \right\|^2$. Resample\cite{song2023solving} first optimizes the objective 
$\| \mathbf{y} - \mathcal{A}(\mathcal{D}(\mathbf{z})) \|_2^2$, and then stochastic resampling from the distribution $\mathcal{N} \left( \sqrt{\bar{\alpha}_t} \hat{\mathbf{z}}_0(\mathbf{y}), \, (1 - \bar{\alpha}_t) \mathbf{I} \right)$ to improve performance.

\subsection{Decoupled Posterior Sampling}

A major limitation of methods that transform the prior into the posterior distribution during the reverse process is the bias introduced in the early stages of sampling, which can be difficult to correct later on. Decoupled Posterior Sampling methods\cite{zhang2024improving, li2024decoupled, alkhouri2024sitcom} address this challenge through a three-step approach: the reverse process, the optimization process, and the forward process. Instead of directly sampling from the posterior distribution, Decoupled Posterior Sampling methods construct a sequence of decoupled noisy samples to mitigate early-stage bias. 

DAPS\cite{zhang2024improving} is a representative work among the Decoupled Posterior Sampling methods. In DAPS, prior noisy samples are first passed through the reverse process to obtain a clear sample \( \hat{\mathbf{x}}_0(\mathbf{x}_t) \). The optimization process then refines the clear sample \( \hat{\mathbf{x}}_0(\mathbf{x}_t) \), producing a sample $\mathbf{x}_{0|\mathbf{y}}$ that considered measurement $\mathbf{y}$. Finally, the forward process reintroduces noise to the sample $\mathbf{x}_{0|\mathbf{y}}$. With noise levels gradually decreasing throughout the stages, the sample in the final stage is noise-free.

\section{Method}
\subsection{Background}
\textbf{Diffusion Models}\\
Diffusion model\cite{ho2020denoising, nichol2021improved, song2019generative, song2020improved, dhariwal2021diffusion, sohl2015deep} consists of two processes: the forward process and the reverse process. In the forward process, the model gradually transforms the data from a clean state to a highly noisy state, eventually reaching a distribution close to random noise. In the reverse process, the model seeks to remove this noise in a step-by-step manner to reconstruct the original data. The forward and reverse processes can be represented using stochastic differential equations, as shown in equations:
\begin{align}
    d\mathbf{x}_t = f(\mathbf{x}_t, t) \, dt + g(t) \, d\mathbf{W}_t,
\end{align}
\begin{align}
    d\mathbf{x}_t = \left( f(\mathbf{x}_t, t) - g^2(t) \nabla_{\mathbf{x}_t} \log p_t(\mathbf{x}_t) \right) dt + g(t) \, d\mathbf{W}_t.
\end{align}

In these equations, \( f(\mathbf{x}_t, t) \) is the drift coefficient, and \( g(t) \) is the diffusion coefficient. As the parameters of the forward process is known, the distribution closed to the random noise get from the forwar process can be used in training the reverse process. Diffusion model is trained to estimate the score function \( \nabla_{\mathbf{x}_t} \log p_t(\mathbf{x}_t) \). Using this estimated score function, the pretrained diffusion model can generate data from random noise through the reverse process.\\
\textbf{Decoupled Posterior Sampling}\\
Recently, a class of methods known as Decoupled Posterior Sampling methods\cite{zhang2024improving, li2024decoupled, alkhouri2024sitcom} have achieved state-of-the-art performance. Decoupled Posterior Sampling methods use a sequence of noisy samples that progressively converge to a clean sample by gradually reducing the added noise. This approach involves three main steps: the reverse process, the optimization process, and the forward process. 

In the reverse process, a denoising step is performed on the noisy sample \( \mathbf{x}_t \). The optimization process then seeks a solution close to the objective \( \arg\min_{\mathbf{x}_0} \frac{1}{2} \| \mathbf{y} - \mathcal{A}(\mathbf{x}_0) \|_2^2 \), where $\mathbf{x}_0$ is the signal we want to reconstruct. Finally, the forward process reintroduces noise into the sample generated from the optimization process. As an example of decoupled posterior sampling methods, we introduce DAPS\cite{zhang2024improving}, which solves the probability flow ODE\cite{karras2022elucidating} in the reverse process, as shown in equation:
\begin{align}
    \text{d}\mathbf{x}_t = -\dot{\sigma}_t \sigma_t \nabla_{\mathbf{x}_t} \log p(\mathbf{x}_t; \sigma_t) \, \text{d}t.
\end{align}
Through the reverse process, we obtain a clear sample \( \hat{\mathbf{x}}_0 \). Assuming the signal \( x_0 \) follows the distribution in equation:
\begin{align}
p(\mathbf{x}_0 \mid \mathbf{x}_t) \approx \mathcal{N}(\mathbf{x}_0 ; \hat{\mathbf{x}}_0(\mathbf{x}_t), r_t^2 \mathbf{I}).
\end{align}

Given that \( p(\mathbf{y} | \mathbf{x}_0) \approx \mathcal{N}(\mathbf{y} ; \mathcal{A}(\mathbf{x}), \beta_t^2 \mathbf{I}) \), we can derive the posterior distribution \( p(\mathbf{x}_0 | \mathbf{x}_t, \mathbf{y}) \) using Bayes’ theorem, as shown in equation:
\begin{align}
p(\mathbf{x}_0 \mid \mathbf{x}_t, y) \propto p(\mathbf{x}_0 \mid \mathbf{x}_t) p(\mathbf{y} \mid \mathbf{x}_0).
\end{align}

DAPS employs Langevin Dynamics\cite{welling2011bayesian} in the optimization process, which samples from the distribution \( p(\mathbf{x}_0 | \mathbf{x}_t, \mathbf{y}) \) starting from \( \hat{\mathbf{x}}_0 \), thereby yielding an approximation \( \mathbf{x}_{0 \mid \mathbf{y}} \) that adheres closely to \( p(\mathbf{x}_0 | \mathbf{x}_t, \mathbf{y}) \).

In the forward process, we generate the next noisy sample \( \mathbf{x}_t \) by sampling from the Gaussian distribution \( \mathcal{N}(\mathbf{x}_t ; \mathbf{x}_{0 \mid \mathbf{y}}, r_{t-1}^2 \mathbf{I}) \).

\subsection{Guided Decoupled Posterior Sampling}
\begin{figure}[h]
    \centering
    \includegraphics[width=0.5\columnwidth]{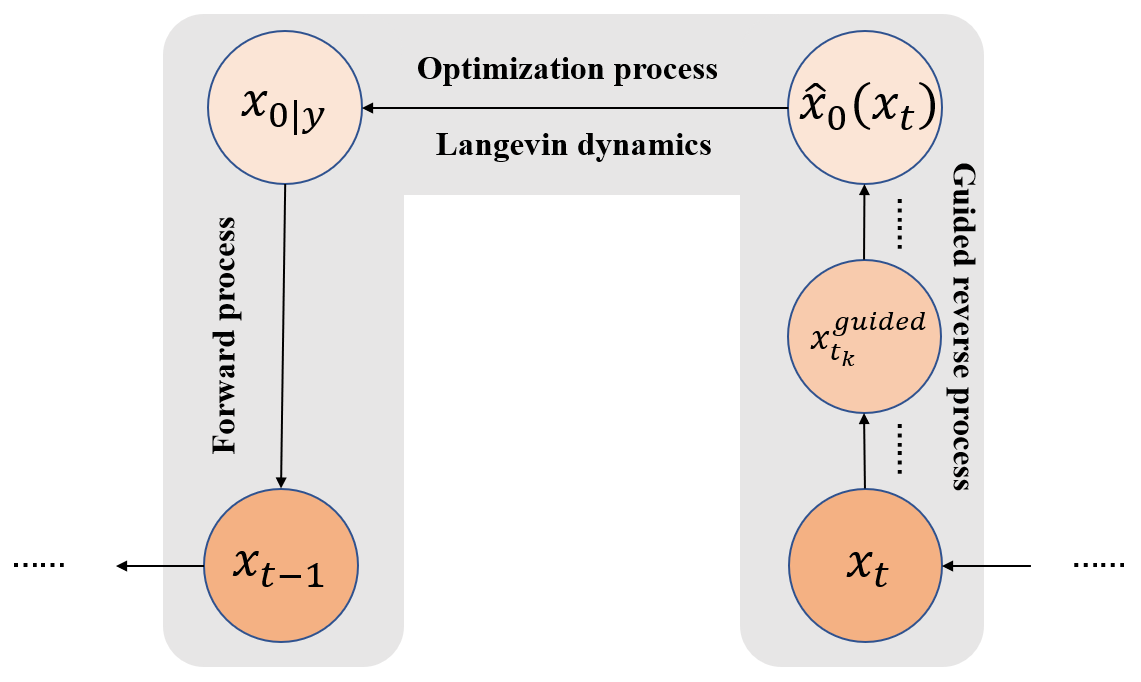}
    \caption{\textbf{Illustrative Diagram of Our Method GDPS.} As shown in the figure, we begin by sending the input \( \mathbf{x}_t \) into the guided reverse process. In this process, we introduce a guidance term to optimize \( \mathbf{x}_{t_k} \) into \( \mathbf{x}_{t_k}^{\text{guided}} \), progressively moving towards a clear sample, \( \hat{\mathbf{x}}_0(\mathbf{x}_t) \). In the optimization process, we apply Langevin dynamics to sample \( \mathbf{x}_{0|\mathbf{y}} \) from the posterior distribution \( p(\mathbf{x}_0 \mid \mathbf{x}_t, \mathbf{y}) \) based on $\hat{\mathbf{x}}_0(\mathbf{x}_t)$. Finally, in the forward process, noise is reintroduced into the sample to obtain the next noisy sample \( \mathbf{x}_{t-1} \), continuing the iterative procedure.}
    \label{fig:example}
\end{figure}
In decoupled posterior sampling methods\cite{zhang2024improving, li2024decoupled, alkhouri2024sitcom}, the reverse process lacks measurement information, leading the optimization process to start from a biased point. Inspired by methods that use Bayesian inference in the reverse process, we aim to balance the independence of the noisy measurement \( \mathbf{y} \) and the clear sample \( \hat{\mathbf{x}}_0 \). To this end, we propose a guided reverse process that generates \( {\mathbf{x}}_{0|\mathbf{y}} \) constrained by data consistency $\| \mathbf{y} - \mathcal{A}(\mathbf{x}) \|_2^2$. As shown in Algorithm \ref{alg: guided-decoupled-posterior-sampling}, instead of generating \( \hat{\mathbf{x}}_0 \) independently of the noisy measurement \( \mathbf{y} \), we incorporate a gradient descent term to correct the state variable following each iteration in solving the probability flow ODE\cite{karras2022elucidating, liu2022flow}:
\begin{align}
    \mathbf{x}_{t}^{\text{guided}} = \mathbf{x}_{t} - \gamma\nabla_{\mathbf{x}_{t}}{||\mathbf{y} - \mathcal{A}(\mathbf{x}_{t})||}^2.
\end{align}
\begin{algorithm}
\caption{GDPS}
\label{alg: guided-decoupled-posterior-sampling}
\small 
\DontPrintSemicolon
  \KwInput{Score model $s_\theta$, measurement $\mathbf{y}$, step size $\gamma$, noisy sample steps $N$, noise schedule $\sigma_{t_i}$, $(t_i)_{i \in \{0, \dots, N\}}$, $(t_k)_{k \in \{0, \dots, n\}}$}
  \KwInitialize{Sample $\mathbf{x}_T \sim \mathcal{N}(0, \sigma_T^2 \mathbf{I})$}
  \For{$i = N, N - 1, \dots, 1$}{
        \For{$k = n, n-1, \ldots, 1$}{
            ${\mathbf{x}}_{{t_{k-1}}} \leftarrow {\mathbf{x}}_{{t_k}} + \dot\sigma(t_k)\sigma(t_k)s_\theta({\mathbf{x}}_{{t_k}}, \sigma(t_{k})\Delta t$. \\
            ${\mathbf{x}}_{{t_{k-1}}} \leftarrow {\mathbf{x}}_{{t_{k-1}}} - \gamma\nabla_{{\mathbf{x}}_{{t_{k-1}}}}{||\mathbf{y} - \mathcal{A}({\mathbf{x}}_{{t_{k-1}}})||}^2$\\
        }
        Sample $\mathbf{x}_{0|\mathbf{y}}$ from the distribution \( p(\mathbf{x}_0 \mid \mathbf{x}_t, \mathbf{y}) \) using Langevin Dynamics with $N_L$ steps;
        
        Sample $\mathbf{x}_{t_i} \sim \mathcal{N}(\mathbf{x}_{0|\mathbf{y}}, \sigma_{t_i}^2 \mathbf{I})$ \;
  }
\KwOutput{$\mathbf{x}_0$}
\end{algorithm}
Here, $\gamma$ represents the step size for the optimization. With the addition of our guidance term, the reverse process now incorporates measurement information \( \mathbf{y} \). This allows for a smooth transition from the noisy sample $\mathbf{x}_t$ to the clear sample $\mathbf{x}_{0|\mathbf{y}}$ in decoupled posterior sampling methods. Our Guided Decoupled Posterior Sampling(GDPS) achieves improved performance over standard decoupled posterior sampling methods\cite{zhang2024improving, li2024decoupled, alkhouri2024sitcom}.

\subsection{Extension to Latent DMs and SITCOM}

Furthermore, our guidance term can be generalized to cases where the reverse process employs latent diffusion models, as in LatentDAPS\cite{zhang2024improving}, or Tweedie’s formula, as in SITCOM\cite{alkhouri2024sitcom}. When applied to latent diffusion models, our guidance term is defined as $\mathbf{z}_{t}^{\text{guided}} = \mathbf{z}_{t} - \gamma\nabla_{\mathbf{z}_{t}}{||\mathbf{y} - \mathcal{A}(\mathcal{D}(\mathbf{z}_{t}))||}^2$, where \( \mathcal{D} \) represents the decoder of the latent diffusion model, and \( \mathbf{z}_{t} \) is the latent representation of \( \mathbf{x}_{t} \), obtained from the encoder \( \mathcal{E} \) as \( \mathbf{z}_{t} = \mathcal{E}(\mathbf{x}_{t}) \). In the case where Tweedie's formula is used, our guidance term simplifies to \( \hat{\mathbf{x}}_0(\mathbf{x}_t)^{\text{guided}} = \hat{\mathbf{x}}_0(\mathbf{x}_t) - \gamma \nabla_{\hat{\mathbf{x}}_0(\mathbf{x}_t)} \| y - \mathcal{A}(\hat{\mathbf{x}}_0(\mathbf{x}_t)) \|^2 \), where \( \hat{\mathbf{x}}_0(\mathbf{x}_t) \) is derived from Tweedie's formula.

We propose Guided Latent Decoupled Annealing Posterior Sampling (G-LatentDAPS) and Guided Step-wise Triple-Consistent Sampling (G-SITCOM) to demonstrate the adaptability of our approach. The corresponding pseudo-code is provided in the Appendix A.1.

\section{Experiments}
\label{sec:experiements}

\begin{table*}[!htbp]
\small 
\centering
\setlength{\tabcolsep}{0pt}
\begin{adjustbox}{width=\linewidth,center}
\begin{tabular*}{\linewidth}{
  @{\extracolsep{\fill}}
  cccccccccccccccc
}
\toprule
& \mc{3}{c}{\textbf{SR} ($\times 4$)} & \mc{3}{c}{\textbf{Inpainting}(box)} & \mc{3}{c}{\textbf{Inpainting}(random)} & \mc{3}{c}{\textbf{Deblur} (Gauss)} & \mc{3}{c}{\textbf{Deblur} (Motion)} \\
\cmidrule{2-4} \cmidrule{5-7} \cmidrule{8-10} \cmidrule{11-13} \cmidrule{14-16}
\textbf{Method} & \footnotesize{PSNR $\uparrow$} & \footnotesize{SSIM $\uparrow$} & \footnotesize{LPIPS $\downarrow$} 
& \footnotesize{PSNR $\uparrow$} & \footnotesize{SSIM $\uparrow$} & \footnotesize{LPIPS $\downarrow$} 
& \footnotesize{PSNR $\uparrow$} & \footnotesize{SSIM $\uparrow$} & \footnotesize{LPIPS $\downarrow$} 
& \footnotesize{PSNR $\uparrow$} & \footnotesize{SSIM $\uparrow$} & \footnotesize{LPIPS $\downarrow$} 
& \footnotesize{PSNR $\uparrow$} & \footnotesize{SSIM $\uparrow$} & \footnotesize{LPIPS $\downarrow$} \\

\toprule
DPS\cite{chung2022diffusion} &  23.77 &	0.669 &	0.281 & 23.19 &	0.803 &	0.194 &	29.80 &	0.856 &	0.183 &	25.08 &	0.704 & 0.242	& 22.44 &	0.617 &	0.287 \\
PSLD\cite{rout2024solving} & 28.05 &	0.785 & 0.248 & 21.45 &	0.689 &	0.334 &	31.02 &	0.870 &	0.202 &	28.32 &	0.803 &	 0.274 & 27.07 & 0.737 & 0.305 \\
Resample\cite{song2023solving} &   23.28 &	0.507 & 0.458 & 19.71 &	0.795 &	0.211 &      29.92 &	0.855 &	0.177 &	26.24 & 0.692 &	0.312 &	25.75 &	0.637 &	0.357 \\
SITCOM\cite{alkhouri2024sitcom} &   27.90 &	0.826 & 0.235 & 24.53 &	0.832 &	0.172 &      30.29 & 0.896 & 0.147 & 27.15 & 0.799 & 0.267 & 29.01 & 0.855 & 0.203 \\
DAPS\cite{zhang2024improving} &   28.97 &	0.816 & 0.182 & 24.75 &	0.830 &	0.135 &      30.79 & 0.865 & 0.130 & 29.07 & 0.815 & 0.178 & 30.99 & 0.863 & 0.133 \\

\midrule
GDPS \small{(ours)} &  \textbf{29.29} & \textbf{0.836} & \textbf{0.169} & \textbf{24.85} & \textbf{0.856}	& \textbf{0.105} & \textbf{31.65} &	\textbf{0.899}	& \textbf{0.091} &	\textbf{29.40} &	\textbf{0.834} &	\textbf{0.165} &	\textbf{31.13} &	\textbf{0.871} &	\textbf{0.130}  \\
\bottomrule
\end{tabular*}
\end{adjustbox}
\caption{Quantitative comparison of the performance of various methods on linear tasks in the \textbf{FFHQ 256x256} dataset. Each task includes noise with \(\sigma = 0.05\). Arrows indicate whether higher (\(\uparrow\)) or lower (\(\downarrow\)) values are better. The best result in each metric is highlighted in \textbf{bold}.}
\label{table:ffhq}
\end{table*}

\begin{table*}[!htbp]
\small 
\centering
\setlength{\tabcolsep}{0pt}
\begin{adjustbox}{width=\linewidth,center}
\begin{tabular*}{\linewidth}{
  @{\extracolsep{\fill}}
  cccccccccccccccc
}
\toprule
& \mc{3}{c}{Phase retrieval} & \mc{3}{c}{Nonlinear deblurring} & \mc{3}{c}{High dynamic range} \\
\cmidrule{2-4} \cmidrule{5-7} \cmidrule{8-10} 
\textbf{Method} & \footnotesize{PSNR $\uparrow$} & \footnotesize{SSIM $\uparrow$} & \footnotesize{LPIPS $\downarrow$} 
& \footnotesize{PSNR $\uparrow$} & \footnotesize{SSIM $\uparrow$} & \footnotesize{LPIPS $\downarrow$} 
& \footnotesize{PSNR $\uparrow$} & \footnotesize{SSIM $\uparrow$} & \footnotesize{LPIPS $\downarrow$} \\

\toprule
DPS\cite{chung2022diffusion} &  16.88 &	0.495 &	0.465 & 22.86 &	0.631 &	0.304 &	23.36 &	0.740 &	0.290 \\
Resample\cite{song2023solving} &   24.32 & 0.620 & 0.371 & 28.53 & 0.793 & 0.217 & 24.82 &	0.837 & 0.177  \\
SITCOM\cite{alkhouri2024sitcom} &   22.05 & 0.653 & 0.386 & 25.49 & 0.725 & 0.315 &      26.21 & 0.855 & 0.202  \\
DAPS\cite{zhang2024improving} &   30.10 & 0.861 & 0.133 & 28.60 & 0.802 & 0.169 &      26.48 & 0.857 & 0.161  \\

\midrule
GDPS \small{(ours)} &  \textbf{31.34} & \textbf{0.892} & \textbf{0.106} & \textbf{28.99} & \textbf{0.824} & \textbf{0.153} & \textbf{27.10} &	\textbf{0.863} & \textbf{0.160}  \\
\bottomrule
\end{tabular*}
\end{adjustbox}
\caption{Quantitative comparison of the performance of various methods on nonlinear tasks in the \textbf{FFHQ 256x256} dataset. Each task includes noise with \(\sigma = 0.05\). Arrows indicate whether higher (\(\uparrow\)) or lower (\(\downarrow\)) values are better. The best result in each metric is highlighted in \textbf{bold}.}
\label{table:ffhq}
\end{table*}

\subsection{Experimental Setup}
\textbf{Tasks}: We evaluate our method on five linear tasks and three nonlinear tasks: (a)for super resolution, we use a bicubic resizer to downscale images from 256x256 to 64x64; (b)for box inpainting, a 128x128 square region is masked in the center of each image; (c)for random inpainting, we randomly remove 70\% of the pixels in the images to create a masking effect; (d)for gaussian deblurring, a gaussian blur with a kernel size of 61x61 and a standard deviation of 3.0 is applied; (e)for motion deblurring, a motion blur with a kernel size of 61x61 and a standard deviation of 0.5 is applied; (f)for high dynamic range, the dynamic range of the images is scaled by a factor of 2.0; (g)for nonlinear deblurring, We follow the default settings provided in \cite{tran2021explore}; (h)for phase retrieval, the oversampling ratio is set to 2.0.

\textbf{Datasets}: We evaluate our method on 100 images from the validation set of the \text{FFHQ 256x256} dataset and \text{ImageNet 256x256} dataset.

\textbf{Pre-trained Diffusion Models}: We use the same pre-trained model provided by \cite{chung2022diffusion}. Additionally, we utilize latent diffusion models trained following the settings in \cite{rombach2022high} with the same setting as \cite{song2023solving}.

\textbf{Baselines}: For linear tasks, we compare our method with DPS\cite{chung2022diffusion}, PSLD\cite{rout2024solving}, Resample\cite{song2023solving}, SITCOM\cite{alkhouri2024sitcom}, and DAPS\cite{zhang2024improving}. For nonlinear tasks, we compare our method with DPS\cite{chung2022diffusion}, Resample\cite{song2023solving}, SITCOM\cite{alkhouri2024sitcom}, and DAPS\cite{zhang2024improving}.

\textbf{Metrics}: To assess the quality of our reconstructed images, we use Peak Signal-to-Noise Ratio (\text{PSNR}), Structural Similarity Index (\text{SSIM})\cite{wang2004image}, and Learned Perceptual Image Patch Similarity (\text{LPIPS})\cite{zhang2018unreasonable} to evaluate our reconstructed images.

\textbf{Running time}: All experiments are conducted on a single NVIDIA GeForce RTX4090 GPU. As shown in Appendix B, the running time of GDPS is compared to DAPS\cite{zhang2024improving}. 

\subsection{Main Results}
As shown in Tables 1 and 2, our GDPS method consistently achieves higher PSNR and SSIM values, along with lower LPIPS scores, compared to DAPS\cite{zhang2024improving}, SITCOM\cite{alkhouri2024sitcom}, Resample\cite{song2023solving}, PSLD\cite{rout2024solving}, and DPS\cite{chung2022diffusion} across both linear and nonlinear tasks. GDPS demonstrates superior performance across all metrics, including PSNR, SSIM, and LPIPS, indicating its effectiveness in reconstructing high-quality images.

\begin{figure*}[!htbp]
     \centering
     \begin{subfigure}[b]{0.9\textwidth}
         \centering
         \includegraphics[width=\textwidth]{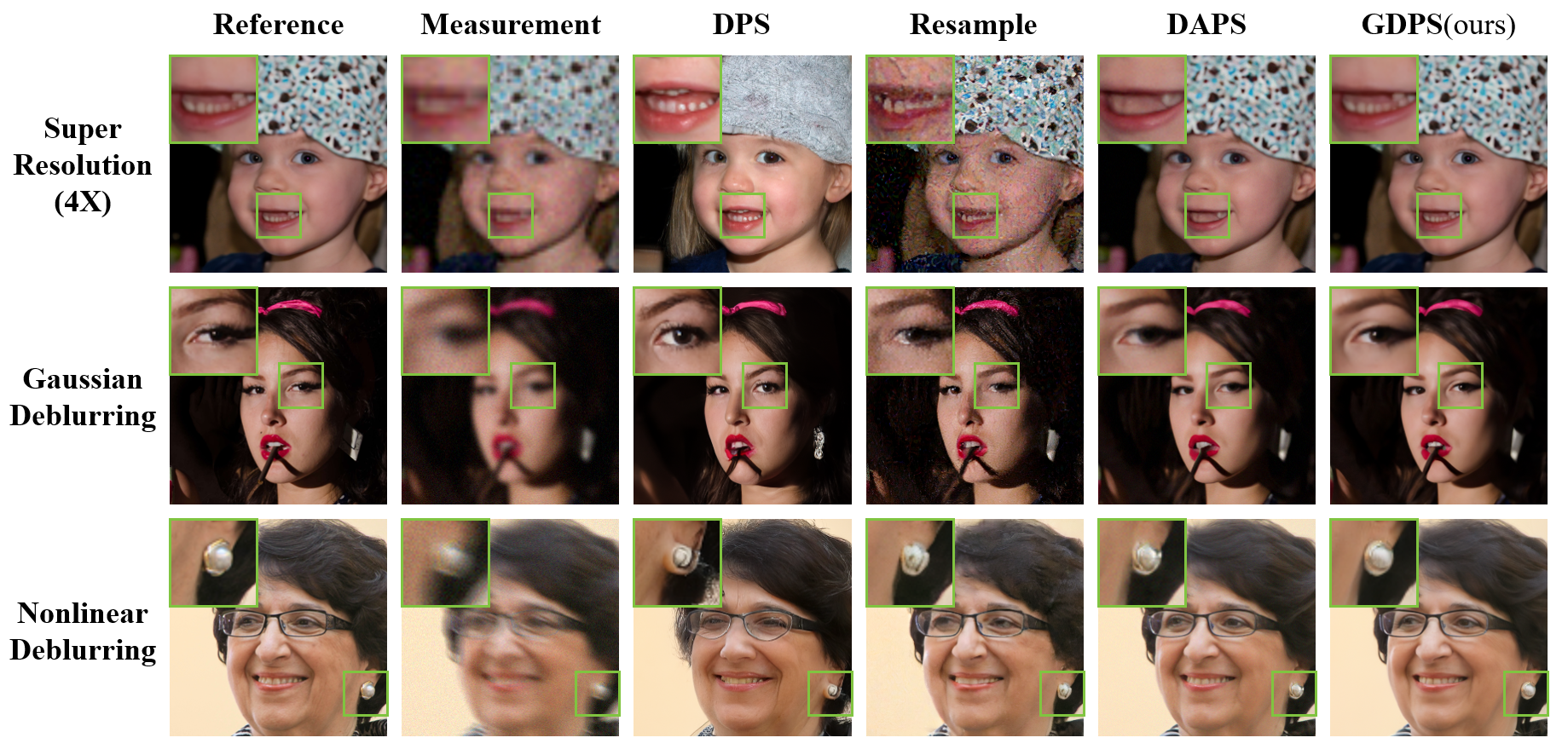}
     \end{subfigure}
        \caption{Representative results from the experiments in the validation set of \textbf{FFHQ 256x256} dataset. Arranged from top to bottom, the taks are: super resolution, gaussian deblurring, and nonlinear deblurring. To highlight areas with quality improvements, we have marked and enlarged specific regions within each image.}
\label{fig-ffhq}
\end{figure*}

\begin{table*}[!htbp]
\small 
\centering
\setlength{\tabcolsep}{0pt}
\begin{adjustbox}{width=\linewidth,center}
\begin{tabular*}{\linewidth}{
  @{\extracolsep{\fill}}
  cccccccccccccccc
}
\toprule
& \mc{3}{c}{\textbf{SR} ($\times 4$)} & \mc{3}{c}{\textbf{Inpainting}(box)} & \mc{3}{c}{\textbf{Inpainting}(random)} & \mc{3}{c}{\textbf{Deblur} (Gauss)} & \mc{3}{c}{\textbf{Deblur} (Motion)} \\
\cmidrule{2-4} \cmidrule{5-7} \cmidrule{8-10} \cmidrule{11-13} \cmidrule{14-16}
\textbf{Method} & \footnotesize{PSNR $\uparrow$} & \footnotesize{SSIM $\uparrow$} & \footnotesize{LPIPS $\downarrow$} 
& \footnotesize{PSNR $\uparrow$} & \footnotesize{SSIM $\uparrow$} & \footnotesize{LPIPS $\downarrow$} 
& \footnotesize{PSNR $\uparrow$} & \footnotesize{SSIM $\uparrow$} & \footnotesize{LPIPS $\downarrow$} 
& \footnotesize{PSNR $\uparrow$} & \footnotesize{SSIM $\uparrow$} & \footnotesize{LPIPS $\downarrow$} 
& \footnotesize{PSNR $\uparrow$} & \footnotesize{SSIM $\uparrow$} & \footnotesize{LPIPS $\downarrow$} \\

\toprule
DPS\cite{chung2022diffusion} &  20.80 &	0.503 &	0.434 & 18.96 &	0.659 &	0.322 &	23.95 &	0.659 &	0.333 &	21.20 &	0.511 & 0.386	& 22.97 &	0.607 &	0.353 \\
PSLD\cite{rout2024solving} & 24.75 & 0.688 & 0.324 & 18.21 & 0.540 & 0.467 & 27.08 & 0.774 &	0.268 &	24.93 &	\textbf{0.698} &	0.333 & 24.07 & 0.632 & 0.368 \\
Resample\cite{song2023solving} &   21.23 &	0.435 & 0.470 & 17.14 &	0.704 &	0.295 &      24.94 &	0.727 &	0.266 &	23.10 & 0.557 &	0.406 &	26.85 &	0.764 &	0.245 \\
SITCOM\cite{alkhouri2024sitcom} &   19.49 &	0.505 & 0.556 & 20.75 &	0.761 &	0.271 &      24.54 & 0.717 & 0.342 & 22.25 & 0.596 & 0.450 & 23.56 & 0.672 & 0.385 \\
DAPS\cite{zhang2024improving} &   24.96 & 0.677 & 0.275 & 21.03 & 0.780 & 0.214 &      26.60 & 0.774 & 0.174 & 24.87 & 0.665 & 0.278 & 27.18 & 0.772 & 0.203 \\

\midrule
GDPS \small{(ours)} &  \textbf{25.10} & \textbf{0.690} & \textbf{0.268} & \textbf{21.26} & \textbf{0.804}	& \textbf{0.190} & \textbf{27.24} &	\textbf{0.809}	& \textbf{0.132} & \textbf{25.13} & 0.684 &	\textbf{0.264} &	\textbf{27.79} & \textbf{0.799} & \textbf{0.176}  \\
\bottomrule
\end{tabular*}
\end{adjustbox}
\caption{Quantitative comparison of the performance of various methods on linear tasks in the \textbf{ImageNet 256x256} dataset. Each task includes noise with \(\sigma = 0.05\). Arrows indicate whether higher (\(\uparrow\)) or lower (\(\downarrow\)) values are better. The best result in each metric is highlighted in \textbf{bold}.}
\label{table:ffhq}
\end{table*}

\begin{table*}[!htbp]
\small 
\centering
\setlength{\tabcolsep}{0pt}
\begin{adjustbox}{width=\linewidth,center}
\begin{tabular*}{\linewidth}{
  @{\extracolsep{\fill}}
  cccccccccccccccc
}
\toprule
& \mc{3}{c}{Phase retrieval} & \mc{3}{c}{Nonlinear deblurring} & \mc{3}{c}{High dynamic range} \\
\cmidrule{2-4} \cmidrule{5-7} \cmidrule{8-10} 
\textbf{Method} & \footnotesize{PSNR $\uparrow$} & \footnotesize{SSIM $\uparrow$} & \footnotesize{LPIPS $\downarrow$} 
& \footnotesize{PSNR $\uparrow$} & \footnotesize{SSIM $\uparrow$} & \footnotesize{LPIPS $\downarrow$} 
& \footnotesize{PSNR $\uparrow$} & \footnotesize{SSIM $\uparrow$} & \footnotesize{LPIPS $\downarrow$} \\

\toprule
DPS\cite{chung2022diffusion} &  15.19 &	0.278 &	0.543 & 20.93 &	0.506 &	0.433 &	14.15 & 0.427 & 0.567 \\
Resample\cite{song2023solving} &   16.23 & 0.279 & 0.556 & 24.69 & 0.670 & 0.316 & 23.76 &	0.767 & 0.254  \\
SITCOM\cite{alkhouri2024sitcom} &   16.85 & 0.382 & 0.589 & 22.77 & 0.619 & 0.404 &      24.91 & 0.850 & 0.206  \\
DAPS\cite{zhang2024improving} &   22.17 & 0.577 & 0.324 & 26.27 & 0.737 & 0.218 &      25.74 & 0.860 & 0.166  \\

\midrule
GDPS \small{(ours)} &  \textbf{22.72} & \textbf{0.580} & \textbf{0.302} & \textbf{26.70} & \textbf{0.759} & \textbf{0.195} & \textbf{25.99} &	\textbf{0.862} & \textbf{0.163}  \\
\bottomrule
\end{tabular*}
\end{adjustbox}
\caption{Quantitative comparison of the performance of various methods on nonlinear tasks in the \textbf{ImageNet 256x256} dataset. Each task includes noise with \(\sigma = 0.05\). Arrows indicate whether higher (\(\uparrow\)) or lower (\(\downarrow\)) values are better. The best result in each metric is highlighted in \textbf{bold}.}
\label{table:ffhq}
\end{table*}

In Figure 3, we illustrate the qualitative performance of GDPS against other methods. In the super resolution task, we highlight and enlarge the child’s teeth in the image. DPS\cite{chung2022diffusion} reconstructs more teeth than present in the reference, while Resample\cite{song2023solving} produces a blurred image, and DAPS\cite{zhang2024improving} reconstructs fewer teeth than the reference. In contrast, GDPS accurately reconstructs the teeth with details closely resembling the reference image, outperforming all other methods. In the Gaussian deblurring task, we focus on the woman’s eye. DPS\cite{chung2022diffusion} produces an eye that appears larger than in the reference, Resample\cite{song2023solving}’s output is blurred, and DAPS\cite{zhang2024improving} yields a partially blurred eye and eyelash. GDPS, however, reconstructs a clear eye and eyelash, closely matching the reference, and delivering the highest visual quality among the methods. For the nonlinear deblurring task, we examine the woman’s earring. DPS\cite{chung2022diffusion} produces a blurred result, Resample\cite{song2023solving} reconstructs an earring that appears embedded in the ear, and DAPS\cite{zhang2024improving} provides a clearer result but still with some embedding. GDPS successfully reconstructs the earring without any embedding and with greater clarity, outperforming the other methods.

\begin{figure*}[!htbp]
     \centering
     \begin{subfigure}[b]{0.9\textwidth}
         \centering
         \includegraphics[width=\textwidth]{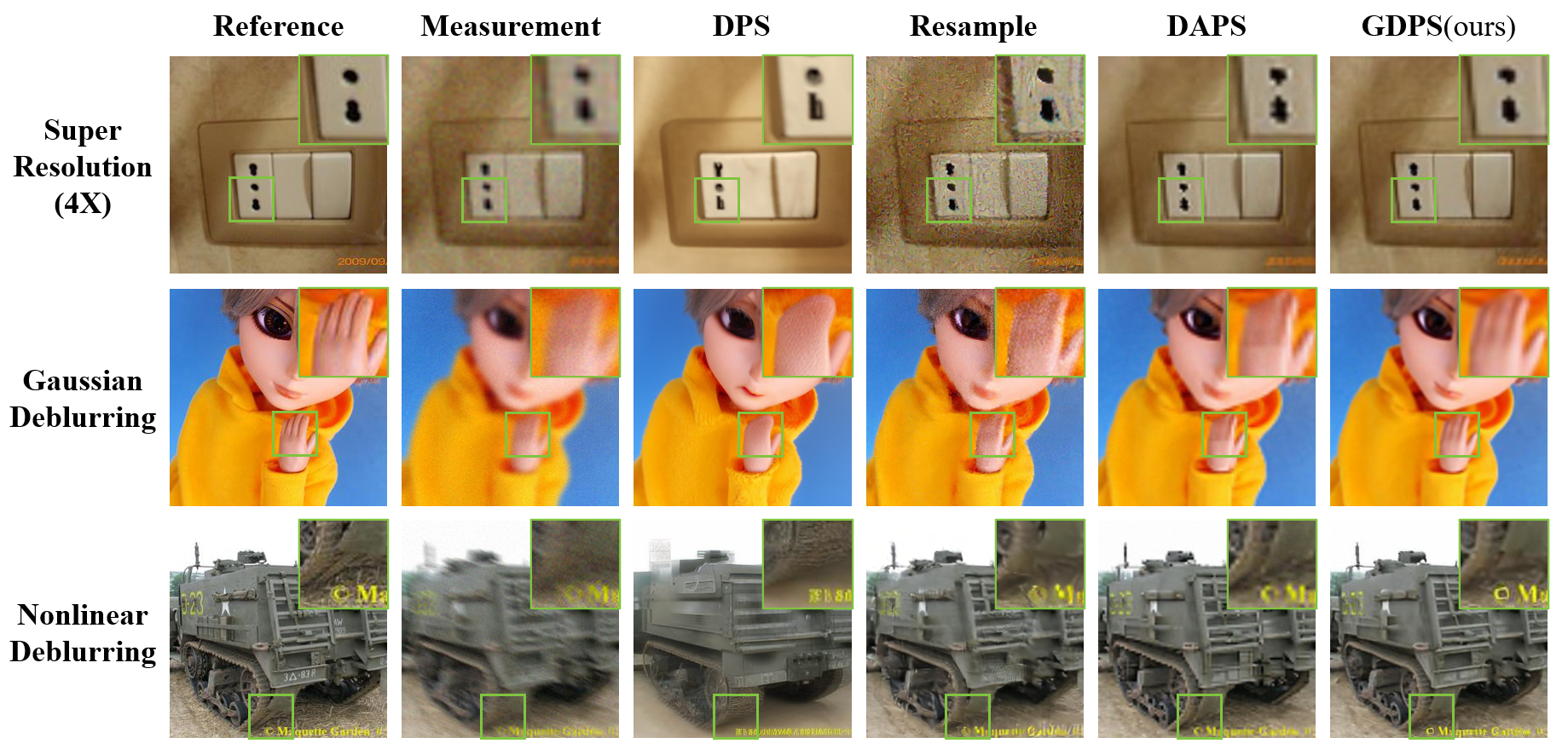}
     \end{subfigure}
        \caption{Representative results from the experiments in the validation set of \textbf{ImageNet 256x256} dataset. Arranged from top to bottom, the taks are: super resolution, gaussian deblurring, and nonlinear deblurring. To highlight areas with quality improvements, we have marked and enlarged specific regions within each image.}
\label{fig-ffhq}
\end{figure*}

In Figure 4, we present further comparisons demonstrating the advantages of GDPS. In the super resolution task, we focus on a socket plug in the image. DPS\cite{chung2022diffusion} reconstructs a plug that appears more squared than the reference, Resample\cite{song2023solving} yields a blurred result, and DAPS\cite{zhang2024improving} introduces aliasing artifacts. GDPS, however, reconstructs the socket plug with high fidelity, closely matching the reference. In the Gaussian deblurring task, we enlarge the cartoon character’s hand for detailed comparison. DPS\cite{chung2022diffusion} reconstructs a hand without distinct fingers, while Resample\cite{song2023solving} produces a hand with blurred, indistinct fingers. DAPS\cite{zhang2024improving} generates a slightly clearer result than Resample, but the fingers remain blurred. In contrast, GDPS reconstructs a clear hand with fingers, surpassing all other methods. Finally, in the nonlinear deblurring task, we examine the tank’s track and letters at the bottom of the image. DPS\cite{chung2022diffusion} reconstructs a track that appears larger than in the reference and produces incorrect letters, Resample\cite{song2023solving} produces a blurred track and letters, and DAPS\cite{zhang2024improving} yields a track with aliasing and blurred letters. GDPS reconstructs both the letters and track with clarity, closely matching the reference and outperforming all other methods in visual fidelity.

\begin{table*}
\small 
\centering
\setlength{\tabcolsep}{0pt}
\begin{adjustbox}{width=\linewidth,center}
\begin{tabular*}{\linewidth}{
  @{\extracolsep{\fill}}
  cccccccccccccccc
}
\toprule
& \mc{3}{c}{\textbf{SR} ($\times 16$)} & \mc{3}{c}{\textbf{Inpainting}($192\times192$)} & \mc{3}{c}{\textbf{Inpainting}(90\%)} \\
\cmidrule{2-4} \cmidrule{5-7} \cmidrule{8-10} 
\textbf{Method} & \footnotesize{PSNR $\uparrow$} & \footnotesize{SSIM $\uparrow$} & \footnotesize{LPIPS $\downarrow$} 
& \footnotesize{PSNR $\uparrow$} & \footnotesize{SSIM $\uparrow$} & \footnotesize{LPIPS $\downarrow$} 
& \footnotesize{PSNR $\uparrow$} & \footnotesize{SSIM $\uparrow$} & \footnotesize{LPIPS $\downarrow$} \\

\toprule
DPS\cite{chung2022diffusion} &   19.08 & 0.469 & 0.425 & 16.41 & 0.593 & 0.345 &      24.95 & 0.708 & 0.294  \\
DAPS\cite{zhang2024improving} &   21.25 & 0.565 & 0.387 & 18.11 & 0.662 & 0.266 &      26.91 & 0.778 & 0.206  \\
GDPS \small{(ours)} &  \textbf{21.96} & \textbf{0.599} & \textbf{0.364} & \textbf{18.37} & \textbf{0.677}	& \textbf{0.249} & \textbf{27.42} &	\textbf{0.811}	& \textbf{0.168}  \\
\bottomrule
\end{tabular*}
\end{adjustbox}
\caption{Quantitative comparison of the performance of various methods on linear challenging tasks in the \textbf{FFHQ 256x256} dataset. Each task includes noise with \(\sigma = 0.05\). Arrows indicate whether higher (\(\uparrow\)) or lower (\(\downarrow\)) values are better. The best result in each metric is highlighted in \textbf{bold}.}
\label{table:ffhq}
\end{table*}

\begin{table*}
\small 
\centering
\setlength{\tabcolsep}{0pt}
\begin{adjustbox}{width=\linewidth,center}
\begin{tabular*}{\linewidth}{
  @{\extracolsep{\fill}}
  ccccccccccccccccc
}
\toprule
& \mc{3}{c}{\textbf{1.5}} & \mc{3}{c}{\textbf{1.0}} & \mc{3}{c}{\textbf{0.5}} & \mc{3}{c}{\textbf{0.0}} \\
\cmidrule{2-4} \cmidrule{5-7} \cmidrule{8-10} \cmidrule{11-13}
\textbf{Method} & \footnotesize{PSNR $\uparrow$} & \footnotesize{SSIM $\uparrow$} & \footnotesize{LPIPS $\downarrow$} 
& \footnotesize{PSNR $\uparrow$} & \footnotesize{SSIM $\uparrow$} & \footnotesize{LPIPS $\downarrow$} 
& \footnotesize{PSNR $\uparrow$} & \footnotesize{SSIM $\uparrow$} & \footnotesize{LPIPS $\downarrow$} 
& \footnotesize{PSNR $\uparrow$} & \footnotesize{SSIM $\uparrow$} & \footnotesize{LPIPS $\downarrow$} \\

\toprule
DPS\cite{chung2022diffusion} & 19.70 & 0.569 & 0.384 & 16.89 & 0.460 & 0.481 & 16.08 & 0.432 & 0.506 & 12.50 & 0.281 & 0.628 \\
DAPS\cite{zhang2024improving} & 30.64 & 0.877 & 0.125 & 29.26 & 0.864 & 0.143 & 27.82 & 0.824 & 0.169 & 18.09 & 0.495 & 0.426 \\
GDPS(ours) & \textbf{31.38} & \textbf{0.887} & \textbf{0.113} & \textbf{30.74} & \textbf{0.881} & \textbf{0.119} & \textbf{28.83} & \textbf{0.847} & \textbf{0.147} & \textbf{19.45} & \textbf{0.546} & \textbf{0.383} \\

\bottomrule
\end{tabular*}
\end{adjustbox}
\caption{Quantitative comparison of the performance of various methods on phase retrieval task in the \textbf{FFHQ 256x256} dataset. Each task includes noise with \(\sigma = 0.05\). Arrows indicate whether higher (\(\uparrow\)) or lower (\(\downarrow\)) values are better. The best result in each metric is highlighted in \textbf{bold}.}
\label{table:oversample}
\end{table*}

\begin{figure*}[!htbp]
     \centering
     \begin{subfigure}[b]{1.0\textwidth}
         \centering
         \includegraphics[width=\textwidth]{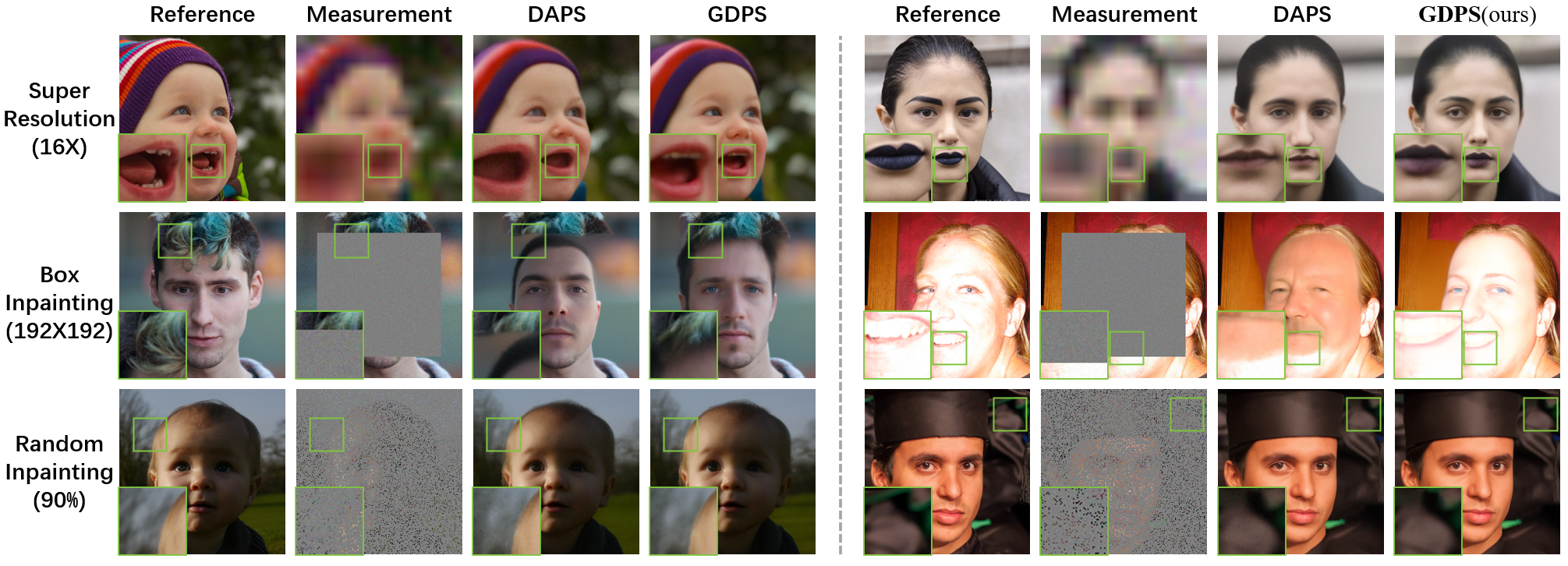}
     \end{subfigure}
        \caption{Representative results from the challenging experiments in the validation set of \textbf{FFHQ 256x256} dataset. Arranged from top to bottom, the taks are: super resolution(16x), box inpainting(192x192), and inpainting(90\%). To highlight areas with quality improvements, we have marked and enlarged specific regions within each image.}
\label{fig-ffhq}
\end{figure*}

\subsection{Challenging tasks}
We evaluate our method on four challenging tasks: (a)for super resolution, we downscale images from 256x256 to 16x16 using bicubic resampling; (b)for box inpainting, a 192x192 box is masked in the center of each image; (c)for random inpainting, we randomly mask 90\% of the pixels in each image.; (d)for phase retrieval, the oversampling factor is varied from 1.5 to 0.0.

As shown in Tables 5 and 6, our GDPS method consistently achieves higher PSNR and SSIM values, along with lower LPIPS scores, compared to the DPS\cite{chung2022diffusion} and DAPS\cite{zhang2024improving} methods across all challenging tasks.

\begin{figure*}[!htbp]
     \centering
     \begin{subfigure}[b]{0.8\textwidth}
         \centering
         \includegraphics[width=\textwidth]{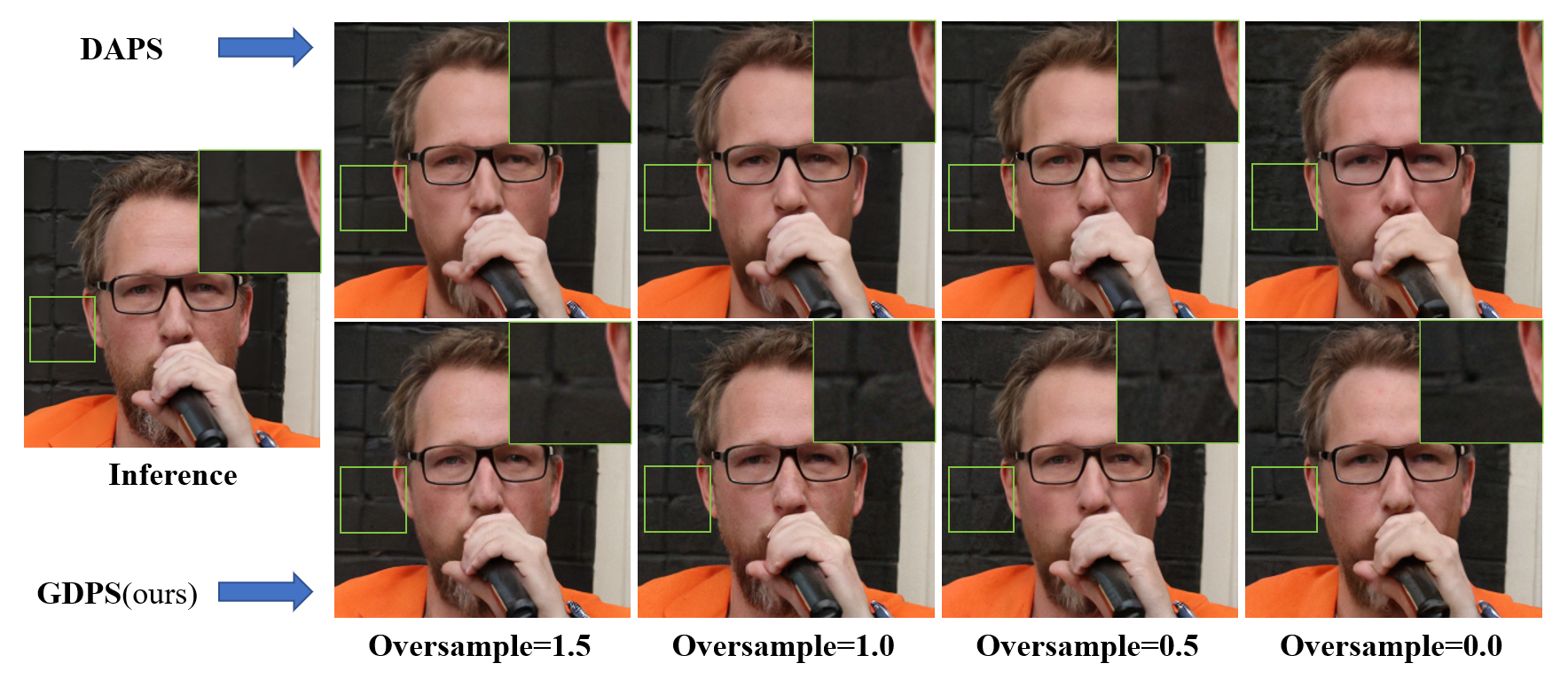}
     \end{subfigure}
        \caption{Representative results from the experiments in the validation set of \textbf{FFHQ 256x256} dataset. Arranged from top to bottom, the methods are: DAPS, and GDPS. From left to right, the oversampling factor varies from 1.5 to 0.0. To highlight areas with quality improvements, we have marked and enlarged specific regions within each image.}
\label{fig-ffhq}
\end{figure*}

Figures 5 and 6 further illustrate the qualitative advantages of our GDPS method over other methods. In the super resolution task, GDPS accurately reconstructs the teeth, whereas DAPS\cite{zhang2024improving} produces a result with an almost toothless mouth in the left part of the images. Additionally, GDPS reconstructs thicker lips that are more similar to the reference than those produced by DAPS\cite{zhang2024improving}. For the box inpainting task, GDPS reconstructs a person with natural-looking hair, while DAPS\cite{zhang2024improving} reconstructs closely cropped hair in the left part of the images. In the right part, GDPS reconstructs a woman with a grinning expression, whereas DAPS\cite{zhang2024improving} reconstructs a man with a closed mouth. In the random inpainting task, GDPS produces clearer images than DAPS\cite{zhang2024improving}, as DAPS\cite{zhang2024improving}’s results appear more blurred. In the phase retrieval task, GDPS reconstructs more detailed textures of the stone wall behind the subject across all oversampling settings. When the oversampling factor is 1.5 and 1.0, GDPS reconstructs the detailed grain of the stone, while DAPS\cite{zhang2024improving} only reconstructs vague lines. At lower oversampling factors (0.5 and 0.0), GDPS still reconstructs complete lines on the stone, whereas DAPS\cite{zhang2024improving} fails to capture these details accurately.

\begin{table}[!htbp]
\centering
\scriptsize 
\setlength{\tabcolsep}{0pt} 
\begin{adjustbox}{width=0.5\linewidth,center} 
\begin{tabular}{@{}ccccccc@{}}
\toprule
\multirow{2}{*}{\textbf{Method}} & \multicolumn{3}{c}{\textbf{SR} ($\times 4$)} & \multicolumn{3}{c}{\textbf{Deblur} (Motion)} \\ 
\cmidrule(lr){2-4} \cmidrule(lr){5-7}
& PSNR $\uparrow$ & SSIM $\uparrow$ & LPIPS $\downarrow$
& PSNR $\uparrow$ & SSIM $\uparrow$ & LPIPS $\downarrow$ \\

\midrule
LatentDAPS\cite{zhang2024improving} & 26.76         & 0.743         & 0.321          & 27.99         & 0.731         & 0.279          \\
G-LatentDAPS & \textbf{27.16}         & \textbf{0.759}         & \textbf{0.303}          & \textbf{28.43}         & \textbf{0.750}         & \textbf{0.262}          \\
\midrule
SITCOM\cite{alkhouri2024sitcom} & 27.90         & 0.826         & 0.235          & 29.01         & 0.855         & 0.203          \\
G-SITCOM & \textbf{28.59} & \textbf{0.855} & \textbf{0.184} & \textbf{29.44} & \textbf{0.870} & \textbf{0.178} \\
\bottomrule
\end{tabular}
\end{adjustbox}
\caption{Quantitative comparison of the performance of various methods on super resolution task and motion deblurring task in the \textbf{FFHQ 256x256} dataset. Each task includes noise with \(\sigma = 0.05\). Arrows indicate whether higher (\(\uparrow\)) or lower (\(\downarrow\)) values are better. The better result in the comparison is highlighted in \textbf{bold}.}
\label{table:comparison}
\end{table}

\subsection{Experiments on Latent DMs and SITCOM}
We assess the effectiveness of our guidance term by applying it into both the LatentDAPS \cite{zhang2024improving} and SITCOM \cite{alkhouri2024sitcom} methods, resulting in our proposed G-LatentDAPS and G-SITCOM variants. A comparative analysis between G-LatentDAPS, G-SITCOM, and their original counterparts, LatentDAPS and SITCOM, is conducted on the \text{FFHQ 256x256} dataset. Experimental results consistently show that incorporating our guidance term enhances performance. A subset of the experimental results on selected tasks is presented in Table 7, with additional results for other tasks provided in the Appendix A.2.

\section{Discussion and Conclusion}
In this paper, we proposed a Guided Decoupled Posterior Sampling (GDPS) method that incorporates data consistency into the reverse process of Decoupled Posterior Sampling, significantly enhancing its effectiveness for solving inverse problems. GDPS achieves state-of-the-art performance across a variety of inverse problem tasks, demonstrating robust performance on both linear and nonlinear problems such as super resolution, gaussian deblurring, and phase retrieval. Notably, our method can be extended to latent diffusion models(LatentDAPS) and SITCOM, showcasing its adaptability. This allows GDPS to deliver improvements across different Decoupled Posterior Sampling methods.

\bibliographystyle{unsrt}  


\begin{thebibliography}{1}

\bibitem{o1986statistical}
Finbarr O'Sullivan.
\newblock A statistical perspective on ill-posed inverse problems.
\newblock {\em Statistical science}, pages 502--518, 1986. JSTOR.

\bibitem{candes2006robust}
Emmanuel J. Candès, Justin Romberg, and Terence Tao.
\newblock Robust uncertainty principles: Exact signal reconstruction from highly incomplete frequency information.
\newblock {\em IEEE Transactions on Information Theory}, 52(2):489--509, 2006. IEEE.

\bibitem{candes2008introduction}
Emmanuel J. Candès and Michael B. Wakin.
\newblock An introduction to compressive sampling.
\newblock {\em IEEE Signal Processing Magazine}, 25(2):21--30, 2008. IEEE.

\bibitem{fazel2008compressed}
Maryam Fazel, Emmanuel Candès, Ben Recht, and Pablo Parrilo.
\newblock Compressed sensing and robust recovery of low rank matrices.
\newblock In {\em 2008 42nd Asilomar Conference on Signals, Systems and Computers}, pages 1043--1047, 2008. IEEE.

\bibitem{foygel2014corrupted}
Rina Foygel and Lester Mackey.
\newblock Corrupted sensing: Novel guarantees for separating structured signals.
\newblock {\em IEEE Transactions on Information Theory}, 60(2):1223--1247, 2014. IEEE.

\bibitem{tang2009performance}
Jie Tang, Brian E. Nett, and Guang-Hong Chen.
\newblock Performance comparison between total variation (TV)-based compressed sensing and statistical iterative reconstruction algorithms.
\newblock {\em Physics in Medicine \& Biology}, 54(19):5781, 2009. IOP Publishing.

\bibitem{ho2020denoising}
Jonathan Ho, Ajay Jain, and Pieter Abbeel.
\newblock Denoising diffusion probabilistic models.
\newblock {\em Advances in Neural Information Processing Systems}, 33:6840--6851, 2020.

\bibitem{nichol2021improved}
Alexander Quinn Nichol and Prafulla Dhariwal.
\newblock Improved denoising diffusion probabilistic models.
\newblock In {\em International Conference on Machine Learning}, pages 8162--8171, 2021. PMLR.

\bibitem{song2019generative}
Yang Song and Stefano Ermon.
\newblock Generative modeling by estimating gradients of the data distribution.
\newblock {\em Advances in Neural Information Processing Systems}, 32, 2019.

\bibitem{song2020improved}
Yang Song and Stefano Ermon.
\newblock Improved techniques for training score-based generative models.
\newblock {\em Advances in Neural Information Processing Systems}, 33:12438--12448, 2020.

\bibitem{dhariwal2021diffusion}
Prafulla Dhariwal and Alexander Nichol.
\newblock Diffusion models beat GANs on image synthesis.
\newblock {\em Advances in Neural Information Processing Systems}, 34:8780--8794, 2021.

\bibitem{sohl2015deep}
Jascha Sohl-Dickstein, Eric Weiss, Niru Maheswaranathan, and Surya Ganguli.
\newblock Deep unsupervised learning using nonequilibrium thermodynamics.
\newblock In {\em International Conference on Machine Learning}, pages 2256--2265, 2015. PMLR.

\bibitem{meng2022quantized}
Xiangming Meng and Yoshiyuki Kabashima.
\newblock Quantized compressed sensing with score-based generative models.
\newblock {\em arXiv preprint arXiv:2211.13006}, 2022.

\bibitem{chung2022diffusion}
Hyungjin Chung, Jeongsol Kim, Michael T. McCann, Marc L. Klasky, and Jong Chul Ye.
\newblock Diffusion posterior sampling for general noisy inverse problems.
\newblock {\em arXiv preprint arXiv:2209.14687}, 2022.

\bibitem{chung2022improving}
Hyungjin Chung, Byeongsu Sim, Dohoon Ryu, and Jong Chul Ye.
\newblock Improving diffusion models for inverse problems using manifold constraints.
\newblock {\em Advances in Neural Information Processing Systems}, 35:25683--25696, 2022.

\bibitem{jalal2021robust}
Ajil Jalal, Marius Arvinte, Giannis Daras, Eric Price, Alexandros G. Dimakis, and Jon Tamir.
\newblock Robust compressed sensing MRI with deep generative priors.
\newblock {\em Advances in Neural Information Processing Systems}, 34:14938--14954, 2021.

\bibitem{jalal2021instance}
Ajil Jalal, Sushrut Karmalkar, Alexandros G. Dimakis, and Eric Price.
\newblock Instance-optimal compressed sensing via posterior sampling.
\newblock {\em arXiv preprint arXiv:2106.11438}, 2021.

\bibitem{kawar2022denoising}
Bahjat Kawar, Michael Elad, Stefano Ermon, and Jiaming Song.
\newblock Denoising diffusion restoration models.
\newblock {\em Advances in Neural Information Processing Systems}, 35:23593--23606, 2022.

\bibitem{kawar2021snips}
Bahjat Kawar, Gregory Vaksman, and Michael Elad.
\newblock SNIPS: Solving noisy inverse problems stochastically.
\newblock {\em Advances in Neural Information Processing Systems}, 34:21757--21769, 2021.

\bibitem{song2023pseudoinverse}
Jiaming Song, Arash Vahdat, Morteza Mardani, and Jan Kautz.
\newblock Pseudoinverse-guided diffusion models for inverse problems.
\newblock In {\em International Conference on Learning Representations}, 2023.

\bibitem{wang2024dmplug}
Hengkang Wang, Xu Zhang, Taihui Li, Yuxiang Wan, Tiancong Chen, and Ju Sun.
\newblock DMPlug: A Plug-in Method for Solving Inverse Problems with Diffusion Models.
\newblock {\em arXiv preprint arXiv:2405.16749}, 2024.

\bibitem{zhang2024improving}
Bingliang Zhang, Wenda Chu, Julius Berner, Chenlin Meng, Anima Anandkumar, and Yang Song.
\newblock Improving diffusion inverse problem solving with decoupled noise annealing.
\newblock {\em arXiv preprint arXiv:2407.01521}, 2024.

\bibitem{li2024decoupled}
Xiang Li, Soo Min Kwon, Ismail R. Alkhouri, Saiprasad Ravishankar, and Qing Qu.
\newblock Decoupled data consistency with diffusion purification for image restoration.
\newblock {\em arXiv preprint arXiv:2403.06054}, 2024.

\bibitem{alkhouri2024sitcom}
Ismail Alkhouri, Shijun Liang, Cheng-Han Huang, Jimmy Dai, Qing Qu, Saiprasad Ravishankar, and Rongrong Wang.
\newblock SITCOM: Step-wise Triple-Consistent Diffusion Sampling for Inverse Problems.
\newblock {\em arXiv preprint arXiv:2410.04479}, 2024.

\bibitem{karras2022elucidating}
Tero Karras, Miika Aittala, Timo Aila, and Samuli Laine.
\newblock Elucidating the design space of diffusion-based generative models.
\newblock {\em Advances in Neural Information Processing Systems}, 35:26565--26577, 2022.

\bibitem{liu2022flow}
Xingchao Liu, Chengyue Gong, and Qiang Liu.
\newblock Flow straight and fast: Learning to generate and transfer data with rectified flow.
\newblock {\em arXiv preprint arXiv:2209.03003}, 2022.

\bibitem{yang2024consistency}
Ling Yang, Zixiang Zhang, Zhilong Zhang, Xingchao Liu, Minkai Xu, Wentao Zhang, Chenlin Meng, Stefano Ermon, and Bin Cui.
\newblock Consistency flow matching: Defining straight flows with velocity consistency.
\newblock {\em arXiv preprint arXiv:2407.02398}, 2024.

\bibitem{boys2023tweedie}
Benjamin Boys, Mark Girolami, Jakiw Pidstrigach, Sebastian Reich, Alan Mosca, and O Deniz Akyildiz.
\newblock Tweedie moment projected diffusions for inverse problems.
\newblock {\em arXiv preprint arXiv:2310.06721}, 2023.

\bibitem{welling2011bayesian}
Max Welling and Yee W. Teh.
\newblock Bayesian learning via stochastic gradient Langevin dynamics.
\newblock In {\em Proceedings of the 28th International Conference on Machine Learning (ICML-11)}, pages 681--

\bibitem{kingma2014adam}
Diederik P. Kingma.
\newblock Adam: A method for stochastic optimization.
\newblock {\em arXiv preprint arXiv:1412.6980}, 2014.

\bibitem{robbins1951stochastic}
Herbert Robbins and Sutton Monro.
\newblock A stochastic approximation method.
\newblock {\em The Annals of Mathematical Statistics}, pages 400--407, 1951.

\bibitem{meng2022diffusion}
Xiangming Meng and Yoshiyuki Kabashima.
\newblock Diffusion model based posterior sampling for noisy linear inverse problems.
\newblock {\em arXiv preprint arXiv:2211.12343}, 2022.

\bibitem{chung2023prompt}
Hyungjin Chung, Jong Chul Ye, Peyman Milanfar, and Mauricio Delbracio.
\newblock Prompt-tuning latent diffusion models for inverse problems.
\newblock {\em arXiv preprint arXiv:2310.01110}, 2023.

\bibitem{rout2024solving}
Litu Rout, Negin Raoof, Giannis Daras, Constantine Caramanis, Alex Dimakis, and Sanjay Shakkottai.
\newblock Solving linear inverse problems provably via posterior sampling with latent diffusion models.
\newblock In {\em Advances in Neural Information Processing Systems}, volume 36, 2024.

\bibitem{song2023solving}
Bowen Song, Soo Min Kwon, Zecheng Zhang, Xinyu Hu, Qing Qu, and Liyue Shen.
\newblock Solving inverse problems with latent diffusion models via hard data consistency.
\newblock {\em arXiv preprint arXiv:2307.08123}, 2023.

\bibitem{karras2019style}
Tero Karras, Samuli Laine, and Timo Aila.
\newblock A style-based generator architecture for generative adversarial networks.
\newblock In {\em Proceedings of the IEEE/CVF Conference on Computer Vision and Pattern Recognition}, pages 4401--4410, 2019.

\bibitem{deng2009imagenet}
Jia Deng, Wei Dong, Richard Socher, Li-Jia Li, Kai Li, and Li Fei-Fei.
\newblock Imagenet: A large-scale hierarchical image database.
\newblock In {\em 2009 IEEE Conference on Computer Vision and Pattern Recognition}, pages 248--255, 2009.

\bibitem{chung2023decomposed}
Hyungjin Chung, Suhyeon Lee, and Jong Chul Ye.
\newblock Decomposed diffusion sampler for accelerating large-scale inverse problems.
\newblock {\em arXiv preprint arXiv:2303.05754}, 2023.

\bibitem{zhu2023denoising}
Yuanzhi Zhu, Kai Zhang, Jingyun Liang, Jiezhang Cao, Bihan Wen, Radu Timofte, and Luc Van Gool.
\newblock Denoising diffusion models for plug-and-play image restoration.
\newblock In {\em Proceedings of the IEEE/CVF Conference on Computer Vision and Pattern Recognition}, pages 1219--1229, 2023.

\bibitem{rout2024beyond}
Litu Rout, Yujia Chen, Abhishek Kumar, Constantine Caramanis, Sanjay Shakkottai, and Wen-Sheng Chu.
\newblock Beyond first-order Tweedie: Solving inverse problems using latent diffusion.
\newblock In {\em Proceedings of the IEEE/CVF Conference on Computer Vision and Pattern Recognition}, pages 9472--9481, 2024.

\bibitem{dou2024diffusion}
Zehao Dou and Yang Song.
\newblock Diffusion posterior sampling for linear inverse problem solving: A filtering perspective.
\newblock In {\em The Twelfth International Conference on Learning Representations}, 2024.

\bibitem{wu2024principled}
Zihui Wu, Yu Sun, Yifan Chen, Bingliang Zhang, Yisong Yue, and Katherine L. Bouman.
\newblock Principled probabilistic imaging using diffusion models as plug-and-play priors.
\newblock {\em arXiv preprint arXiv:2405.18782}, 2024.

\bibitem{sun2024provable}
Yu Sun, Zihui Wu, Yifan Chen, Berthy T. Feng, and Katherine L. Bouman.
\newblock Provable probabilistic imaging using score-based generative priors.
\newblock {\em IEEE Transactions on Computational Imaging}, 2024.

\bibitem{cardoso2023monte}
Gabriel Cardoso, Sylvain Le Corff, Eric Moulines, and others.
\newblock Monte Carlo guided denoising diffusion models for Bayesian linear inverse problems.
\newblock In {\em The Twelfth International Conference on Learning Representations}, 2023.

\bibitem{kim2021noise2score}
Kwanyoung Kim and Jong Chul Ye.
\newblock Noise2score: Tweedie’s approach to self-supervised image denoising without clean images.
\newblock In {\em Advances in Neural Information Processing Systems}, volume 34, pages 864--874, 2021.

\bibitem{efron2011tweedie}
Bradley Efron.
\newblock Tweedie’s formula and selection bias.
\newblock {\em Journal of the American Statistical Association}, volume 106, number 496, pages 1602--1614, 2011.

\bibitem{tran2021explore}
Phong Tran, Anh Tuan Tran, Quynh Phung, and Minh Hoai.
\newblock Explore image deblurring via encoded blur kernel space.
\newblock In {\em Proceedings of the IEEE/CVF Conference on Computer Vision and Pattern Recognition}, pages 11956--11965, 2021.

\bibitem{rombach2022high}
Robin Rombach, Andreas Blattmann, Dominik Lorenz, Patrick Esser, and Björn Ommer.
\newblock High-resolution image synthesis with latent diffusion models.
\newblock In {\em Proceedings of the IEEE/CVF Conference on Computer Vision and Pattern Recognition}, pages 10684--10695, 2022.

\bibitem{zhang2018unreasonable}
Richard Zhang, Phillip Isola, Alexei A. Efros, Eli Shechtman, and Oliver Wang.
\newblock The unreasonable effectiveness of deep features as a perceptual metric.
\newblock In {\em Proceedings of the IEEE Conference on Computer Vision and Pattern Recognition}, pages 586--595, 2018.

\bibitem{wang2004image}
Zhou Wang, Alan C. Bovik, Hamid R. Sheikh, and Eero P. Simoncelli.
\newblock Image quality assessment: From error visibility to structural similarity.
\newblock {\em IEEE Transactions on Image Processing}, volume 13, number 4, pages 600--612, 2004.

\bibitem{chung2024direct}
Hyungjin Chung, Jeongsol Kim, and Jong Chul Ye.
\newblock Direct diffusion bridge using data consistency for inverse problems.
\newblock In {\em Advances in Neural Information Processing Systems}, volume 36, 2024.

\bibitem{alkhouri2024diffusion}
Ismail Alkhouri, Shijun Liang, Rongrong Wang, Qing Qu, and Saiprasad Ravishankar.
\newblock Diffusion-based adversarial purification for robust deep MRI reconstruction.
\newblock In {\em ICASSP 2024-2024 IEEE International Conference on Acoustics, Speech and Signal Processing (ICASSP)}, pages 12841--12845, 2024.

\bibitem{huang2005inverse}
Sixun Huang, Jie Xiang, Huadong Du, and Xiaoqun Cao.
\newblock Inverse problems in atmospheric science and their application.
\newblock In {\em Journal of Physics: Conference Series}, volume 12, number 1, pages 45, 2005.

\bibitem{murata2023gibbsddrm}
Naoki Murata, Koichi Saito, Chieh-Hsin Lai, Yuhta Takida, Toshimitsu Uesaka, Yuki Mitsufuji, and Stefano Ermon.
\newblock Gibbsddrm: A partially collapsed Gibbs sampler for solving blind inverse problems with denoising diffusion restoration.
\newblock In {\em International Conference on Machine Learning}, pages 25501--25522, 2023.

\bibitem{rozet2024learning}
François Rozet, Gérôme Andry, François Lanusse, and Gilles Louppe.
\newblock Learning Diffusion Priors from Observations by Expectation Maximization.
\newblock {\em arXiv preprint arXiv:2405.13712}, 2024.

\end{thebibliography}

\newpage
\appendix

\section{Extension to Latent DMs and SITCOM}
As discussed in the main paper, our method can be extended to LatentDAPS and SITCOM, resulting in G-LatentDAPS and G-SITCOM. In this section, we provide the corresponding pseudo-code and present additional experimental results.
\subsection{pseudo-code}
\begin{algorithm}
\caption{G-LatentDAPS}
\label{alg: decoupled-annealing-posterior-sampling}
\small 
\DontPrintSemicolon
\KwInput{Latent space score model $\mathbf{s}_\theta$, measurement $\mathbf{y}$, step size $\gamma$, noisy sample steps $N$, noise schedule $\sigma_{t_i}$, $(t_i)_{i \in \{0, \dots, N\}}$, $(t_k)_{k \in \{0, \dots, n\}}$, encoder $\mathcal{E}$, decoder $\mathcal{D}$}
\KwInitialize{Sample $\mathbf{z}_T \sim \mathcal{N}(0, \sigma_T^2 \mathbf{I})$}
\For{$i = N, N - 1, \dots, 1$}{
    \textbf{Pixel space Langevin dynamics:}\\
    \For{$k = n, n-1, \ldots, 1$}{
            ${\mathbf{z}}_{{t_{k-1}}} \leftarrow {\mathbf{z}}_{{t_k}} + \dot\sigma(t_k)\sigma(t_k)s_\theta({\mathbf{z}}_{{t_k}}, \sigma(t_{k}))\Delta t$. \\
        }
    $\mathbf{{x}}_{t_0} \gets \mathcal{D}(\mathbf{{z}}_{t_0})$. \;
    Sample $\mathbf{x}_{0|\mathbf{y}}$ from the distribution \( p(\mathbf{x}_{0} \mid \mathbf{x}_{t_0}, \mathbf{y}) \) using Langevin Dynamics with $N_L$ steps;\\
    $\mathbf{z}_{0|\mathbf{y}} \gets \mathcal{E}(\mathbf{x}_{0|\mathbf{y}})$. \;
    \textbf{Latent space Langevin dynamics:}\\
    \For{$k = n, n-1, \ldots, 1$}{
            ${\mathbf{z}}_{{t_{k-1}}} \leftarrow {\mathbf{z}}_{{t_k}} + \dot\sigma(t_k)\sigma(t_k)s_\theta({\mathbf{z}}_{{t_k}}, \sigma(t_{k}))\Delta t$. \\
            ${\mathbf{z}}_{{t_{k-1}}} \leftarrow {\mathbf{z}}_{{t_{k-1}}} - \gamma\nabla_{{\mathbf{z}}_{{t_{k-1}}}}{||\mathbf{y} - \mathcal{A}(\mathcal{D}({\mathbf{z}}_{{t_{k-1}}}))||}^2$.\\
        }
    Sample $\mathbf{z}_{0|\mathbf{y}}$ from the distribution \( p(\mathbf{z}_0 \mid \mathbf{z}_{t_0}, \mathbf{y}, \mathcal{D}) \) using Langevin Dynamics with $N_L$ steps;
    Sample $\mathbf{z}_{t_i} \sim \mathcal{N}(\mathbf{z}_{0|\mathbf{y}}, \sigma_{t_i}^2 \mathbf{I})$ \;
}
\KwOutput{$\mathcal{D}(\mathbf{z}_0)$}
\end{algorithm}

As shown in Algorithm 2, G-LatentDAPS operates in two stages. In the first stage, the latent variable \( \mathbf{z}_t \) is decoded into \( \mathbf{x}_t \), and Langevin dynamics is performed in the pixel space, as described by the following equations:
\begin{align}
\mathbf{x}_0^{(j+1)} = \mathbf{x}_0^{(j)} - \eta \cdot \mathbf{grad} + \sqrt{2\eta}\bm{\epsilon}_j,
\end{align}

\begin{align}
\mathbf{grad} = \nabla_{\mathbf{x}_0^{(j)}} \left(\frac{\|\mathbf{x}_0^{(j)} - \mathcal{D}(\mathbf{z}_{t_0})\|^2}{2r_{t_0}^2} 
+ \frac{\|A(\mathbf{x}) - \mathbf{y}\|^2}{2\beta_y^2} 
\right).
\end{align}

In the second stage of G-LatentDAPS, we introduce our guidance term 
$\mathbf{z}_{t}^{\text{guided}} = \mathbf{z}_{t} - \gamma \nabla_{\mathbf{z}_{t}} \| \mathbf{y} - \mathcal{A}(\mathcal{D}(\mathbf{z}_{t})) \|^2$
and perform Langevin dynamics in the latent space, as described by the following equations:

\begin{align}
\mathbf{z}_0^{(j+1)} = \mathbf{z}_0^{(j)} - \eta \cdot \mathbf{grad} + \sqrt{2\eta}\bm{\epsilon}_j,
\end{align}

\begin{align}
\mathbf{grad} = \nabla_{\mathbf{z}_0^{(j)}} \left( 
\frac{\|\mathbf{z}_0^{(j)} - \mathbf{z}_{t_0}\|^2}{2r_{t_0}^2} 
+ \frac{\|A(\mathcal{D}(\mathbf{z}_0^{(j)})) - \mathbf{y}\|^2}{2\beta_y^2} 
\right).
\end{align}

For G-SITCOM, we define a distance function F($\mathbf{x}$, $\mathbf{y}$) as follows:
\begin{align}
        F(\mathbf{x}, \mathbf{y}) = \left\|\mathcal{A}\left(\frac{1}{\sqrt{\bar{\alpha}_t}} 
        \left[\mathbf{x} - \sqrt{1 - \bar{\alpha}_t} \epsilon_\theta(\mathbf{x}, t)\right]\right) - \mathbf{y}\right\|_2^2.
\end{align}
This function first computes $\hat{\mathbf{x}}_0(\mathbf{x})$ from $\mathbf{x}$ using Tweedie's formula, and then calculates the distance between $\mathcal{A}(\hat{\mathbf{x}}_0(\mathbf{x}))$ and $\mathbf{y}$.

\begin{algorithm}
\caption{G-SITCOM}
\label{alg: guided-decoupled-posterior-sampling}
\small 
\DontPrintSemicolon
  \KwInput{Score model $\epsilon_\theta(\cdot, \cdot)$, distance function $F(\cdot, \cdot)$, measurement $\mathbf{y}$, number of diffusion steps $N$, DM noise schedule $\bar{\alpha}_i$ for $i \in \{1, \ldots, N\}$, regularization parameter $\lambda$}
  \KwInitialize{Sample $\mathbf{x}_T \sim \mathcal{N}(0, \mathbf{I})$}
  \For{$t = N, N - 1, \dots, 1$}{
        $\hat{\mathbf{v}}_t \leftarrow \arg\min_{\mathbf{v}_t'}\left( 
        F(\mathbf{v}_t', \mathbf{y}) + \lambda \left\|\mathbf{x}_t - \mathbf{v}_t'\right\|_2^2\right).$

        $\hat{\mathbf{x}}_0' \leftarrow 
        \frac{1}{\sqrt{\bar{\alpha}_t}} 
        \left[\hat{\mathbf{v}}_t - \sqrt{1 - \bar{\alpha}_t} \epsilon_\theta(\hat{\mathbf{v}}_t, t)\right].$

        $\hat{\mathbf{x}}_0' \leftarrow \hat{\mathbf{x}}_0' - \gamma\nabla_{\hat{\mathbf{x}}_0'}{||\mathbf{y} - \mathcal{A}(\hat{\mathbf{x}}_0')||}^2$.\\

        $\mathbf{x}_{t-1} \leftarrow \sqrt{\bar{\alpha}_{t-1}} \hat{\mathbf{x}}_0' 
        + \sqrt{1 - \bar{\alpha}_{t-1}} \eta_t, \quad \eta_t \sim \mathcal{N}(0, \mathbf{I}).$
  }
\KwOutput{$\mathbf{x}_0$}
\end{algorithm}
As shown in Algorithm 3, our guidance term is updated as $\hat{\mathbf{x}}_0' = \hat{\mathbf{x}}_0' - \gamma\nabla_{\hat{\mathbf{x}}_0'}{||\mathbf{y} - \mathcal{A}(\hat{\mathbf{x}}_0')||}^2$, where \( \hat{\mathbf{x}}_0(\mathbf{x}_t) \) is computed using Tweedie's formula.

\subsection{Experiments on the FFHQ}
We compare our G-LatentDAPS and G-SITCOM methods with LatentDAPS and SITCOM on 100 images from the validation set of the \text{FFHQ 256x256} dataset. The evaluation is conducted on five linear tasks and three nonlinear tasks, with all settings identical to the standard conditions described in the main paper. As shown in Table 8 and Table 9, G-LatentDAPS outperforms LatentDAPS, and G-SITCOM demonstrates superior performance compared to SITCOM across all selected tasks.

\begin{table*}[!htbp]
\small 
\centering
\setlength{\tabcolsep}{0pt}
\begin{adjustbox}{width=\linewidth,center}
\begin{tabular*}{\linewidth}{
  @{\extracolsep{\fill}}
  cccccccccccccccc
}
\toprule
& \mc{3}{c}{\textbf{SR} ($\times 4$)} & \mc{3}{c}{\textbf{Inpainting}(box)} & \mc{3}{c}{\textbf{Inpainting}(random)} & \mc{3}{c}{\textbf{Deblur} (Gauss)} & \mc{3}{c}{\textbf{Deblur} (Motion)} \\
\cmidrule{2-4} \cmidrule{5-7} \cmidrule{8-10} \cmidrule{11-13} \cmidrule{14-16}
\textbf{Method} & \footnotesize{PSNR $\uparrow$} & \footnotesize{SSIM $\uparrow$} & \footnotesize{LPIPS $\downarrow$} 
& \footnotesize{PSNR $\uparrow$} & \footnotesize{SSIM $\uparrow$} & \footnotesize{LPIPS $\downarrow$} 
& \footnotesize{PSNR $\uparrow$} & \footnotesize{SSIM $\uparrow$} & \footnotesize{LPIPS $\downarrow$} 
& \footnotesize{PSNR $\uparrow$} & \footnotesize{SSIM $\uparrow$} & \footnotesize{LPIPS $\downarrow$} 
& \footnotesize{PSNR $\uparrow$} & \footnotesize{SSIM $\uparrow$} & \footnotesize{LPIPS $\downarrow$} \\

\toprule
LatentDAPS\cite{zhang2024improving} & 26.76 & 0.743 & 0.321 & 24.13 &	0.825 &	0.200 &  29.71 & 0.851 & 0.177 & 27.72 & 0.786 & 0.256 & 27.99 & 0.731 & 0.279 \\
G-LatentDAPS\small{(ours)} &   \textbf{27.16} &	\textbf{0.759} & \textbf{0.303} & \textbf{24.29} &	\textbf{0.827} &	\textbf{0.197} & \textbf{29.80} & \textbf{0.856} & \textbf{0.165} & \textbf{27.87} & \textbf{0.789} & \textbf{0.252} & \textbf{28.43} & \textbf{0.750} & \textbf{0.262} \\

\midrule
SITCOM\cite{alkhouri2024sitcom} &  27.90 & 0.826 & 0.235 & 24.53 & 0.832	& 0.172 & 30.29 & 0.896	& 0.147 & 27.15 & 0.799 &	0.267 &	29.01 & 0.855 & 0.203  \\
G-SITCOM\small{(ours)} &  \textbf{28.59} & \textbf{0.855} & \textbf{0.184} & \textbf{24.70} & \textbf{0.859}	& \textbf{0.110} & \textbf{30.68} &	\textbf{0.907}	& \textbf{0.111} & \textbf{27.51} & \textbf{0.814} &	\textbf{0.247} &	\textbf{29.44} & \textbf{0.870} & \textbf{0.178}  \\
\bottomrule
\end{tabular*}
\end{adjustbox}
\caption{Quantitative comparison of the performance of LatentDAPS, SITCOM, G-LatentDAPS and G-SITCOM methods on linear tasks in the \textbf{FFHQ 256x256} dataset. Each task includes noise with \(\sigma = 0.05\). Arrows indicate whether higher (\(\uparrow\)) or lower (\(\downarrow\)) values are better. The better result in the comparison is highlighted in \textbf{bold}.}
\label{table:ffhq}
\end{table*}

\begin{table*}[!htbp]
\small 
\centering
\setlength{\tabcolsep}{0pt}
\begin{adjustbox}{width=\linewidth,center}
\begin{tabular*}{\linewidth}{
  @{\extracolsep{\fill}}
  cccccccccccccccc
}
\toprule
& \mc{3}{c}{Phase retrieval} & \mc{3}{c}{Nonlinear deblurring} & \mc{3}{c}{High dynamic range} \\
\cmidrule{2-4} \cmidrule{5-7} \cmidrule{8-10} 
\textbf{Method} & \footnotesize{PSNR $\uparrow$} & \footnotesize{SSIM $\uparrow$} & \footnotesize{LPIPS $\downarrow$} 
& \footnotesize{PSNR $\uparrow$} & \footnotesize{SSIM $\uparrow$} & \footnotesize{LPIPS $\downarrow$} 
& \footnotesize{PSNR $\uparrow$} & \footnotesize{SSIM $\uparrow$} & \footnotesize{LPIPS $\downarrow$} \\

\toprule
LatentDAPS\cite{zhang2024improving} &   28.30 & 0.814 & 0.238 & 27.57 & 0.758 & 0.264 & \textbf{24.82} & 0.811 & 0.250  \\
G-LatentDAPS\small{(ours)} &   \textbf{28.63} & \textbf{0.819} & \textbf{0.230} & \textbf{27.82} & \textbf{0.766} & \textbf{0.247} &      \textbf{24.82} & \textbf{0.815} & \textbf{0.246}  \\

\midrule
SITCOM\cite{alkhouri2024sitcom} &  22.05 & 0.653 & 0.386 & 25.49 & 0.725 & 0.315 & 26.21 &	0.855 & 0.202  \\
G-SITCOM\small{(ours)} &  \textbf{25.88} & \textbf{0.772} & \textbf{0.277} & \textbf{25.74} & \textbf{0.735} & \textbf{0.313} & \textbf{26.25} &	\textbf{0.870} & \textbf{0.172}  \\
\bottomrule
\end{tabular*}
\end{adjustbox}
\caption{Quantitative comparison of the performance of LatentDAPS, SITCOM, G-LatentDAPS and G-SITCOM methods on nonlinear tasks in the \textbf{FFHQ 256x256} dataset. Each task includes noise with \(\sigma = 0.05\). Arrows indicate whether higher (\(\uparrow\)) or lower (\(\downarrow\)) values are better. The better result in the comparison is highlighted in \textbf{bold}.}
\label{table:ffhq}
\end{table*}

\section{Running time}
We compare the running time of our GDPS method with that of DAPS on 5 images from the validation set of the \text{FFHQ 256x256} dataset. The evaluation is conducted on five linear tasks, with all settings consistent with the standard conditions described in the main paper. All experiments are performed on a single NVIDIA GeForce RTX 4090 GPU. As shown in Table 10, the running time of GDPS is comparable to that of DAPS \cite{zhang2024improving}.
\begin{table}[h!]
\centering
\begin{tabular}{ccccccc}
\toprule
\textbf{Method} & \textbf{SR} & \textbf{IB} & \textbf{IR} & \textbf{GD} & \textbf{MD} \\ \midrule
DAPS\cite{zhang2024improving}    & 62.72 & 43.59 & 45.14 & 50.81 & 51.50 \\ 
GDPS\small{(ours)}  & {64.58} & {47.34} & {48.18} & {54.59} & {55.28} \\ \midrule
difference    & {1.86}  & {3.75}  & {3.04}  & {3.78}  & {3.78} \\ \bottomrule
\end{tabular}
\caption{Quantitative comparison of the running time of DAPS and GDPS methods on linear tasks in the \textbf{FFHQ 256x256} dataset. Each task includes noise with \(\sigma = 0.05\). From the left to right, the tasks are: SR: super resolution, IB: box inpainting, IR: random inpainting, GD: gaussian deblurring, MD: motion deblurring.}
\label{tab:comparison}
\end{table}

\section{Experimental Details}
\subsection{Hyperparameters}
The step sizes $\gamma$ of GDPS, G-LatentDAPS, G-SITCOM for selected linear tasks are shown in Table 11. The step sizes $\gamma$ of GDPS, G-LatentDAPS, G-SITCOM for selected nonlinear tasks are shown in Table 12.
\begin{table}[h!]
\centering
\begin{tabular}{ccccccc}
\toprule
\textbf{Method} & \textbf{SR} & \textbf{IB} & \textbf{IR} & \textbf{GD} & \textbf{MD} \\ \midrule
GDPS    & 2.0 & 3.0 & 5.0 & 10.0 & 8.0 \\ 
G-LatentDAPS  & 3.0 & 1.0 & 6.0 & 2.0 & 3.0 \\
G-SITCOM    & 7.0 & 0.5 & 0.5 & 1.0 & 0.5 \\ \bottomrule
\end{tabular}
\caption{The step sizes $\gamma$ of GDPS, G-LatentDAPS, G-SITCOM methods on linear tasks in the \textbf{FFHQ 256x256} dataset are provided. Each task includes noise with \(\sigma = 0.05\). From the left to right, the tasks are: SR: super resolution, IB: box inpainting, IR: random inpainting, GD: gaussian deblurring, MD: motion deblurring.}
\label{tab:comparison}
\end{table}
\begin{table}[h!]
\centering
\begin{tabular}{ccccc}
\toprule
\textbf{Method} & \textbf{PR} & \textbf{ND} & \textbf{HDR} \\ \midrule
GDPS    & 7.0 & 1.0 & 2.0 \\ 
G-LatentDAPS    & 8.0 & 3.0 & 1.0 \\
G-SITCOM    & 4.0  & 0.1  & 0.1 \\ \bottomrule
\end{tabular}
\caption{The step sizes $\gamma$ of GDPS, G-LatentDAPS, G-SITCOM methods on nonlinear tasks in the \textbf{FFHQ 256x256} dataset are provided. Each task includes noise with \(\sigma = 0.05\). From left to right, the tasks are: PR: phase retrieval, ND: nonlinear deblurring, HDR: high dynamic range.}
\label{tab:comparison}
\end{table}

\subsection{Sampling steps}
For our GDPS method, we choose the noisy sample step as N=400 and the reverse step as n=10 for nonlinear deblurring task and phase retrieval task, the noisy sample step as N=200 and the reverse step as n=5 for other tasks. Additionally, we compare the performance of GDPS and DAPS across different noisy sample step and reverse step configurations in the super resolution task. As shown in Table 13, the experimental results demonstrate that GDPS consistently outperforms DAPS under various settings.
\begin{table*}[!htbp]
\small 
\centering
\setlength{\tabcolsep}{0pt}
\begin{adjustbox}{width=\linewidth,center}
\begin{tabular*}{\linewidth}{
  @{\extracolsep{\fill}}
  cccccccccccccccc
}
\toprule
& \mc{3}{c}{N=100, n=5} & \mc{3}{c}{N=200, n=5} & \mc{3}{c}{N=300, n=5} \\
\cmidrule{2-4} \cmidrule{5-7} \cmidrule{8-10} 
\textbf{Method} & \footnotesize{PSNR $\uparrow$} & \footnotesize{SSIM $\uparrow$} & \footnotesize{LPIPS $\downarrow$} 
& \footnotesize{PSNR $\uparrow$} & \footnotesize{SSIM $\uparrow$} & \footnotesize{LPIPS $\downarrow$} 
& \footnotesize{PSNR $\uparrow$} & \footnotesize{SSIM $\uparrow$} & \footnotesize{LPIPS $\downarrow$} \\

\midrule
\textbf{DAPS}\cite{zhang2024improving} & 28.90 & 0.977 & 0.184 & 28.97 & 0.978 & 0.182 & 28.97 & 0.979 & 0.183 \\
\textbf{GDPS}\small{(ours)} & \textbf{29.27} & \textbf{0.983} & \textbf{0.168} & \textbf{29.31} & \textbf{0.983} & \textbf{0.169} & \textbf{29.30} & \textbf{0.983} & \textbf{0.171} \\

\midrule
& \mc{3}{c}{N=100, n=10} & \mc{3}{c}{N=200, n=10} & \mc{3}{c}{N=300, n=10} \\
\cmidrule{2-4} \cmidrule{5-7} \cmidrule{8-10} 
\textbf{Method} & \footnotesize{PSNR $\uparrow$} & \footnotesize{SSIM $\uparrow$} & \footnotesize{LPIPS $\downarrow$} 
& \footnotesize{PSNR $\uparrow$} & \footnotesize{SSIM $\uparrow$} & \footnotesize{LPIPS $\downarrow$} 
& \footnotesize{PSNR $\uparrow$} & \footnotesize{SSIM $\uparrow$} & \footnotesize{LPIPS $\downarrow$} \\

\midrule
\textbf{DAPS}\cite{zhang2024improving} & 28.64 & 0.976 & 0.180 & 28.74 & 0.978 & 0.175 & 28.76 & 0.978 & 0.174 \\
\textbf{GDPS}\small{(ours)} & \textbf{28.99} & \textbf{0.981} & \textbf{0.168} & \textbf{29.09} & \textbf{0.982} & \textbf{0.164} & \textbf{29.12} & \textbf{0.982} & \textbf{0.163} \\
\bottomrule
\end{tabular*}
\end{adjustbox}
\caption{Comparison of DAPS and GDPS methods under different parameter settings N and n on the super resolution task. Arrows indicate whether higher (\(\uparrow\)) or lower (\(\downarrow\)) values are better. The better result in the comparison is highlighted in \textbf{bold}.}
\label{table:comparison}
\end{table*}

\section{Baseline Details}
\textbf{DPS}: We adopt the same hyperparameter settings as specified in the DPS\cite{chung2022diffusion}. For the high dynamic range task, we set 
$\xi_i = \frac{0.2}{\|\mathbf{y} - \mathcal{A}(\hat{\mathbf{x}}_0(\mathbf{x}_i))\|}$.

\textbf{PSLD}: We follow the same hyperparameter settings as outlined in the PSLD\cite{rout2024solving}. Specifically, we use the Stable Diffusion v1-5 model, as mentioned in the provided code, which yields improved results.

\textbf{Resample}: We utilize the same hyperparameter settings as described in the Resample\cite{song2023solving}.

\textbf{SITCOM}: We use the same sampling steps \( N \), optimization steps \( K \), and regularization parameter \( \lambda \) as detailed in the SITCOM\cite{alkhouri2024sitcom}. For the stopping criterion \( \delta \), we follow the configuration provided by the authors on their GitHub page: 
\url{https://github.com/sjames40/SITCOM/issues/1}.

\textbf{DAPS}: We apply the same hyperparameter settings as specified in the DAPS\cite{zhang2024improving}.

\textbf{LatentDAPS}: We adhere to the hyperparameter settings outlined in the DAPS\cite{zhang2024improving}.

\section{More results}
We present additional results of GDPS on the \textbf{FFHQ 256x256} and \textbf{ImageNet 256x256} datasets. Figure 7 illustrates the performance of GDPS on linear tasks within the \textbf{FFHQ 256x256} dataset, while Figure 8 highlights the results on nonlinear tasks in the same dataset. Similarly, Figure 9 displays the results of GDPS on linear tasks within the \textbf{ImageNet 256x256} dataset, and Figure 10 showcases its performance on nonlinear tasks in the \textbf{ImageNet 256x256} dataset. Figure 11 presents results on several challenging tasks within the \textbf{FFHQ 256x256} dataset.

\newpage

\begin{figure*}[!htbp]
     \centering
     \begin{subfigure}[b]{0.6\textwidth}
         \centering
         \includegraphics[width=\textwidth]{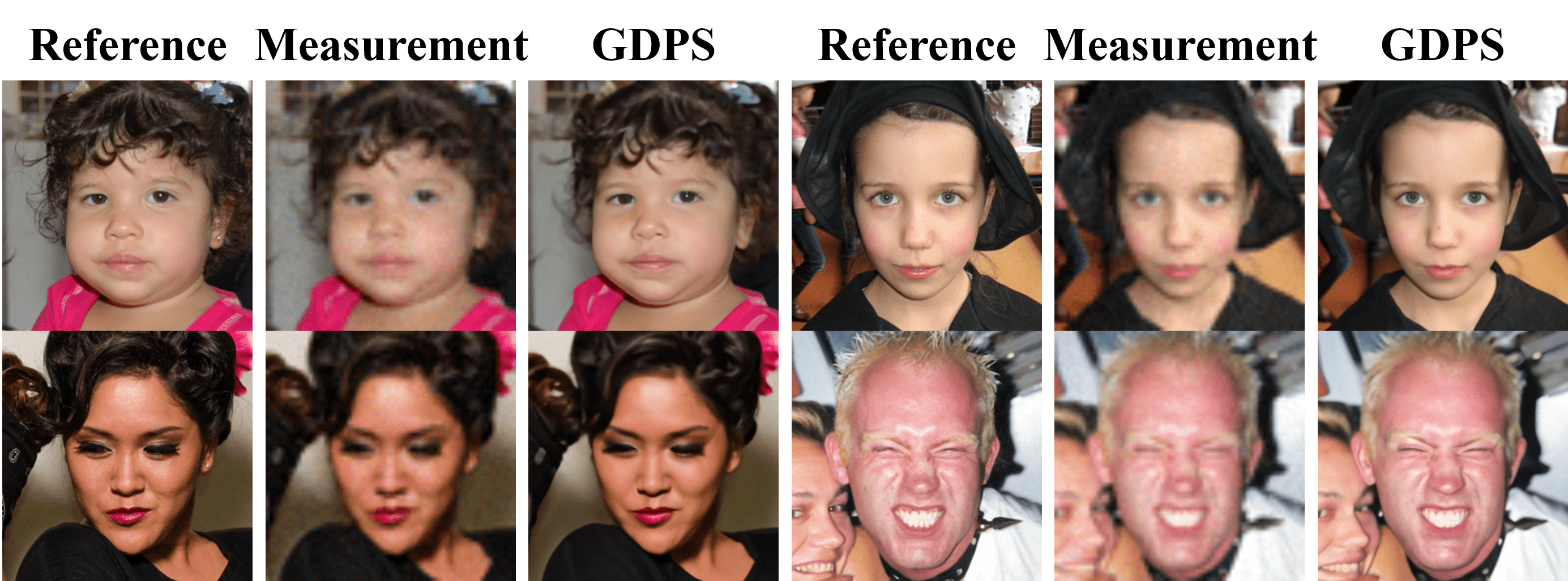}
         \caption{Super Resolution(4x)}
     \end{subfigure}
     \begin{subfigure}[b]{0.6\textwidth}
         \centering
         \includegraphics[width=\textwidth]{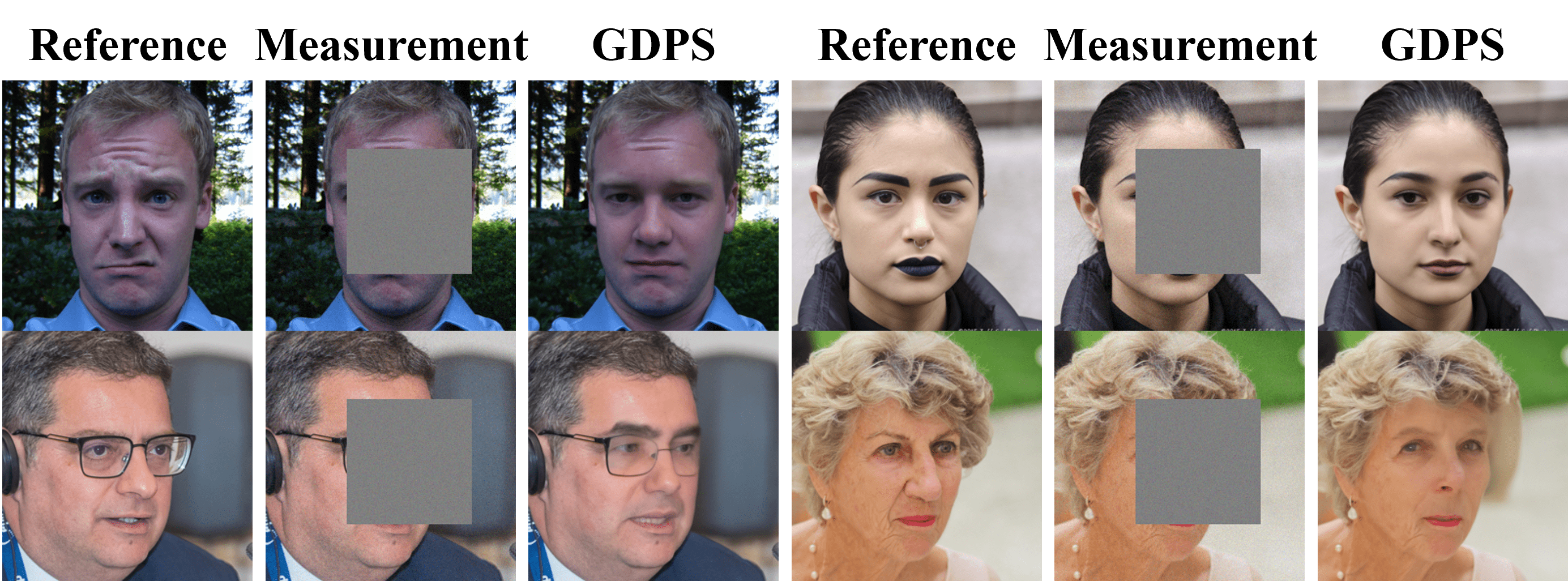}
         \caption{Box Inpainting(128x128)}
     \end{subfigure}
     \begin{subfigure}[b]{0.6\textwidth}
         \centering
         \includegraphics[width=\textwidth]{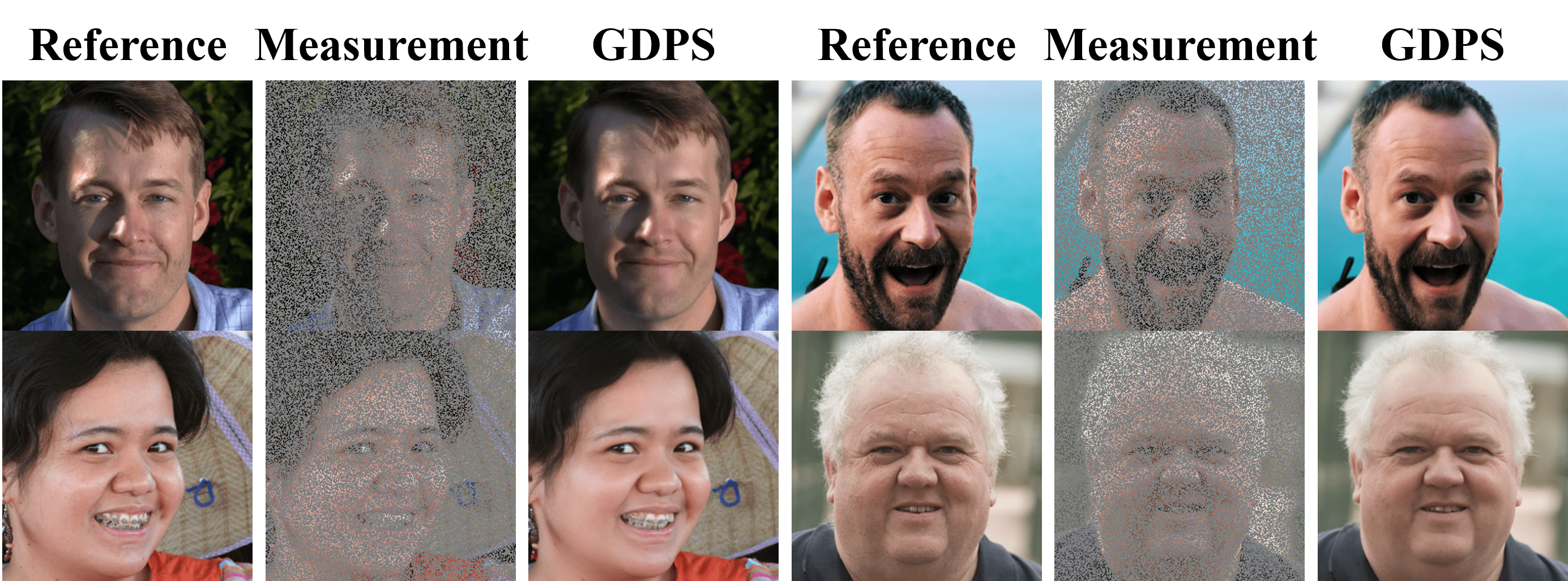}
         \caption{Random Inpainting(70\%)}
     \end{subfigure}
     \begin{subfigure}[b]{0.6\textwidth}
         \centering
         \includegraphics[width=\textwidth]{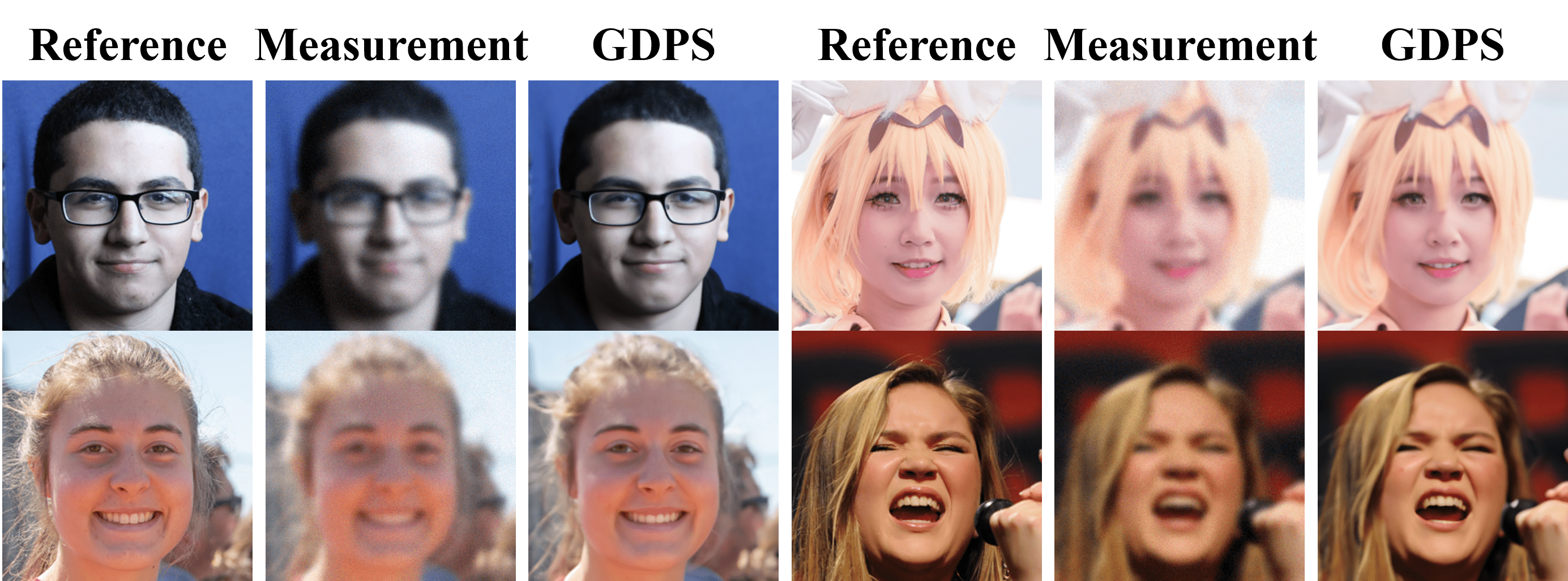}
         \caption{Gaussian Deblurring}
     \end{subfigure}
     \begin{subfigure}[b]{0.6\textwidth}
         \centering
         \includegraphics[width=\textwidth]{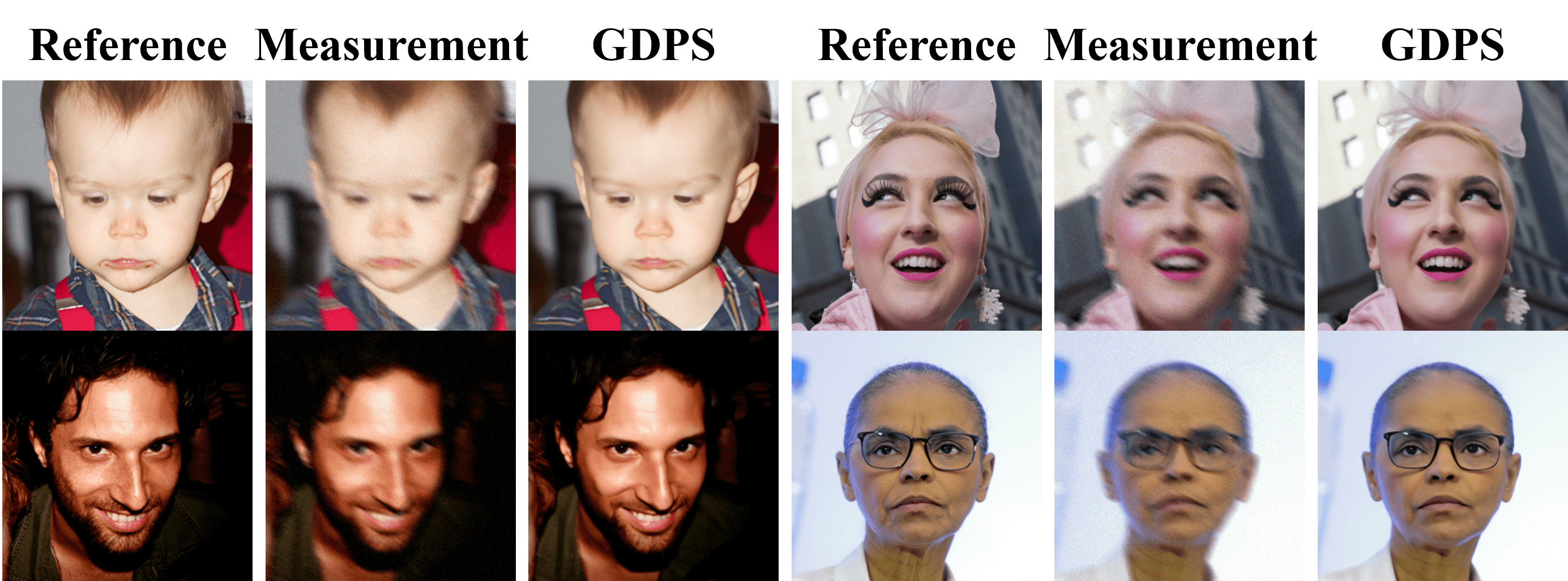}
         \caption{Motion Deblurring}
     \end{subfigure}
        \caption{More results from the experiments in the validation set of \textbf{FFHQ 256x256} dataset. Arranged from top to bottom, the taks are: super resolution, box inpainting, random inpainting, gaussian deblurring and motion deblurring.}
\label{fig-ffhq}
\end{figure*}

\newpage

\begin{figure*}[!htbp]
     \centering
     \begin{subfigure}[b]{0.6\textwidth}
         \centering
         \includegraphics[width=\textwidth]{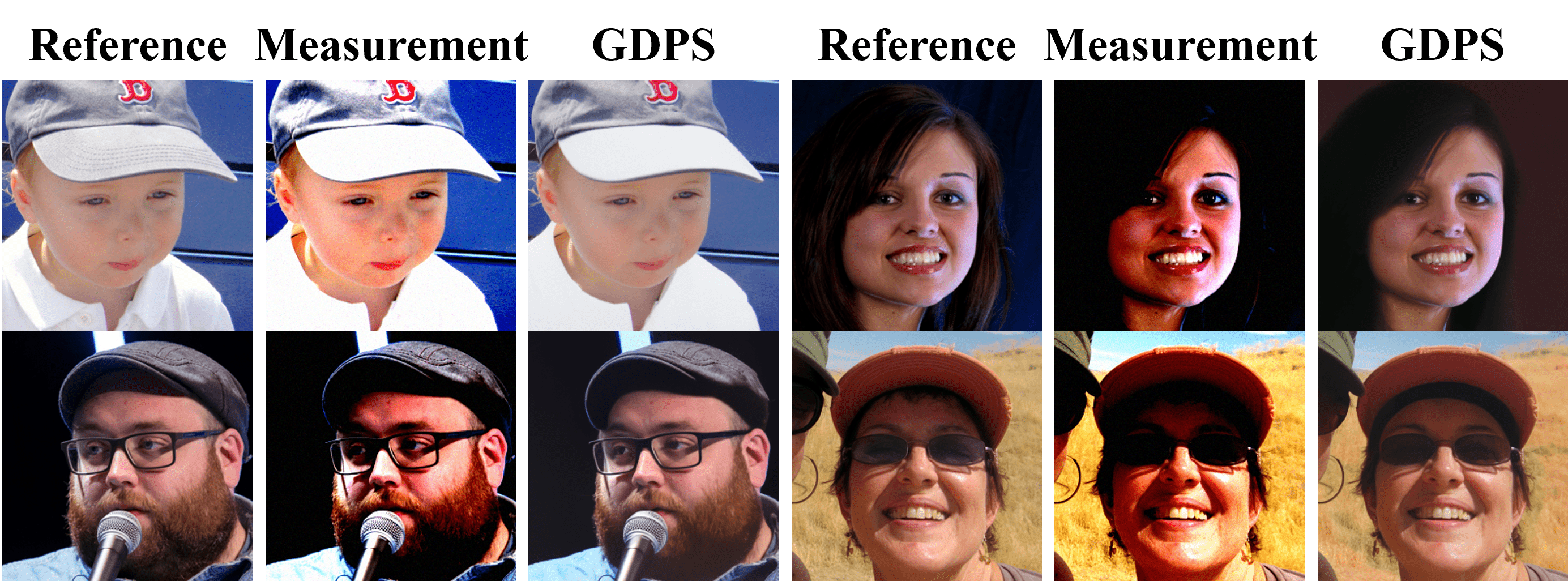}
         \caption{High Dynamic Range}
     \end{subfigure}
     \begin{subfigure}[b]{0.6\textwidth}
         \centering
         \includegraphics[width=\textwidth]{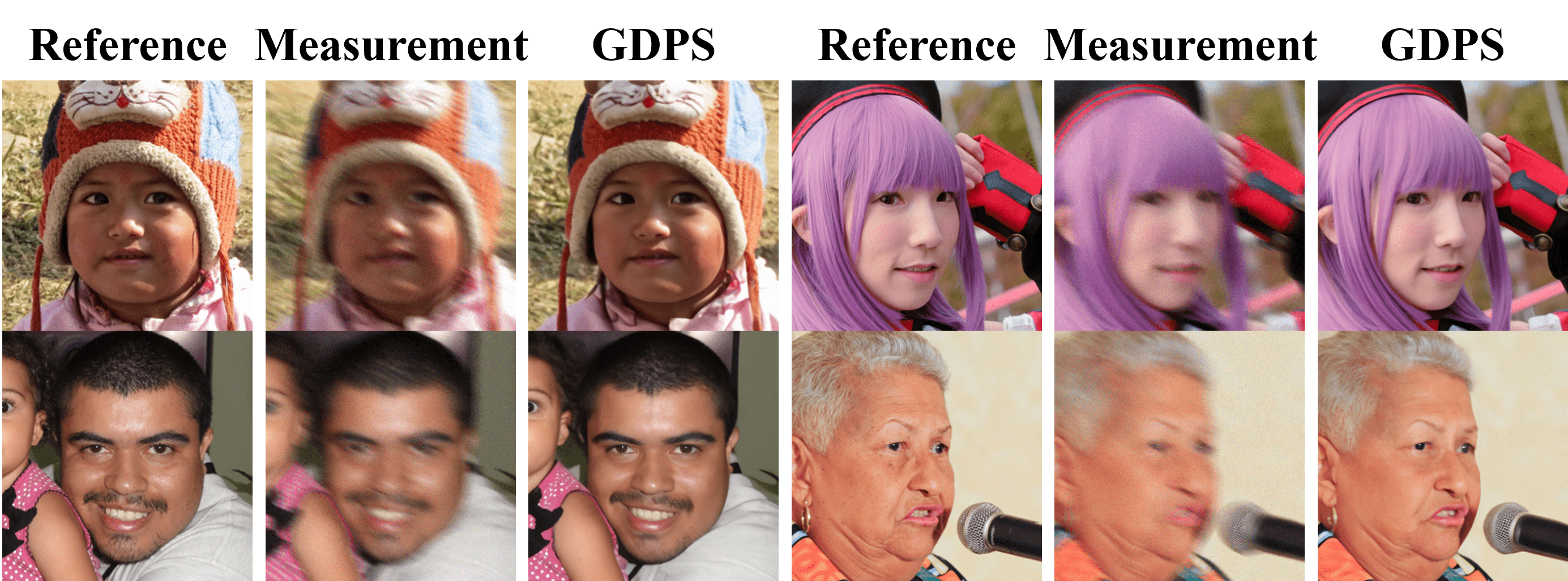}
         \caption{Nonlinear Deblurring}
     \end{subfigure}
     \begin{subfigure}[b]{0.6\textwidth}
         \centering
         \includegraphics[width=\textwidth]{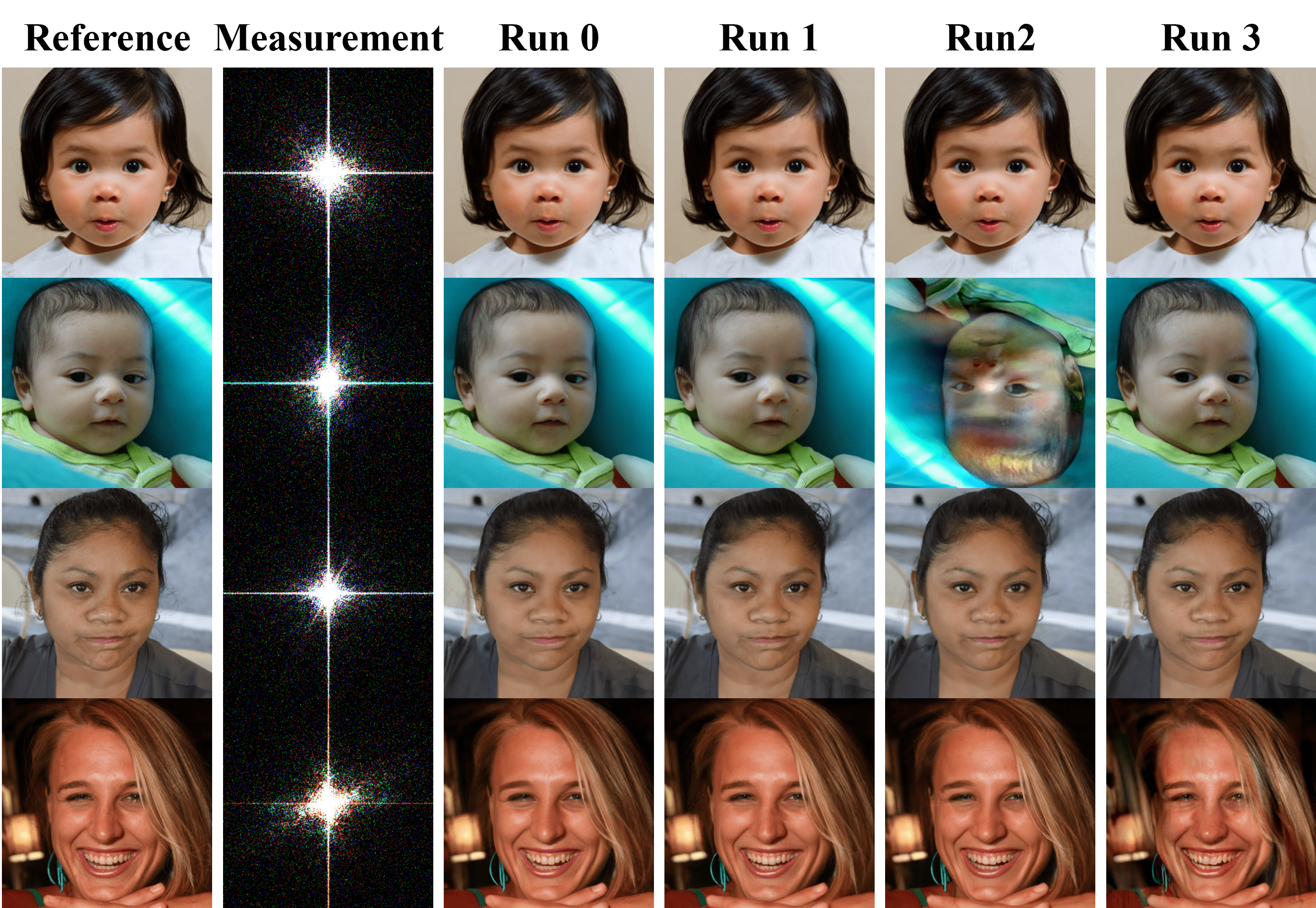}
         \caption{Phase Retrieval}
     \end{subfigure}
        \caption{More results from the experiments in the validation set of \textbf{FFHQ 256x256} dataset. Arranged from top to bottom, the taks are: high dynamic range, nonlinear deblurring and phase retrieval.}
\label{fig-ffhq}
\end{figure*}

\newpage

\begin{figure*}[!htbp]
     \centering
     \begin{subfigure}[b]{0.6\textwidth}
         \centering
         \includegraphics[width=\textwidth]{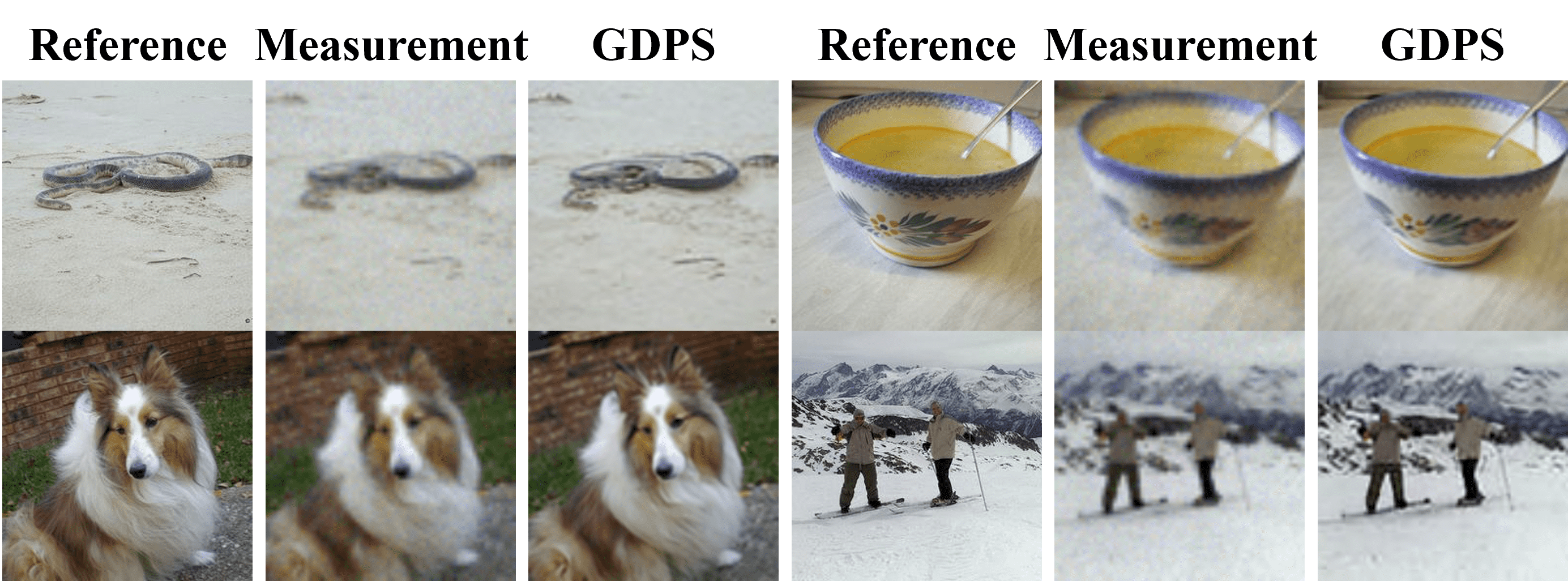}
         \caption{Super Resolution(4x)}
     \end{subfigure}
     \begin{subfigure}[b]{0.6\textwidth}
         \centering
         \includegraphics[width=\textwidth]{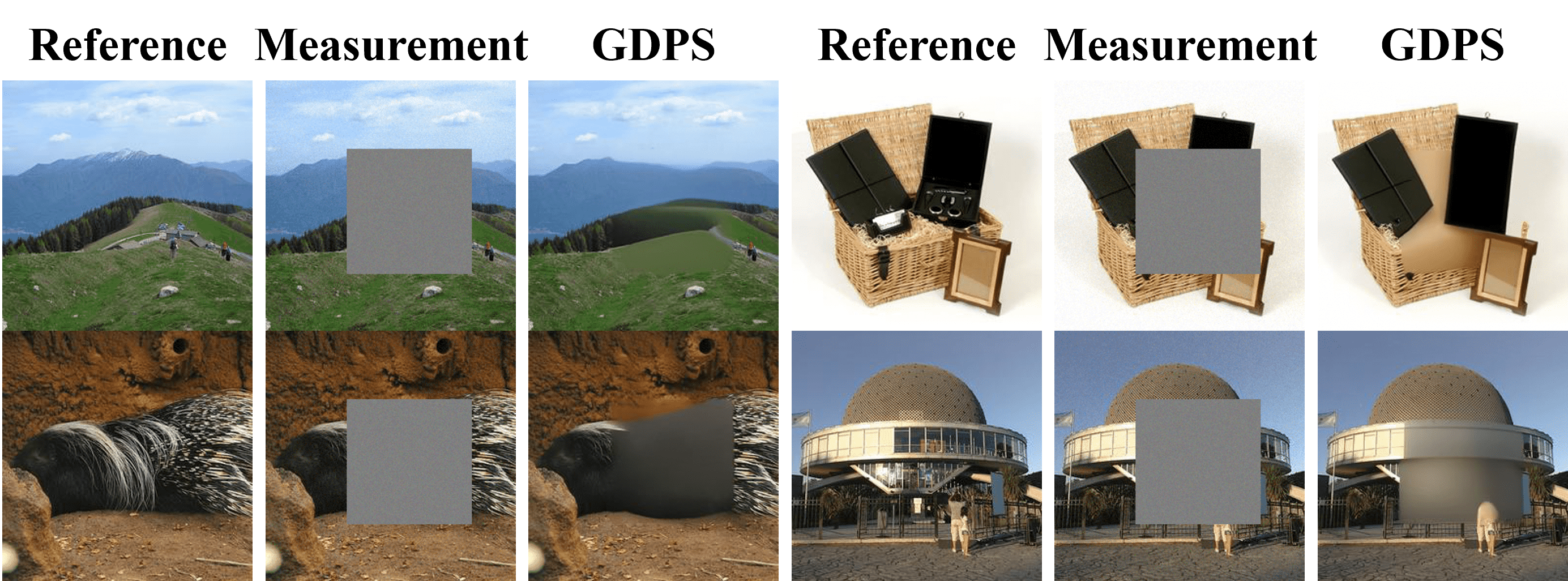}
         \caption{Box Inpainting(128x128)}
     \end{subfigure}
     \begin{subfigure}[b]{0.6\textwidth}
         \centering
         \includegraphics[width=\textwidth]{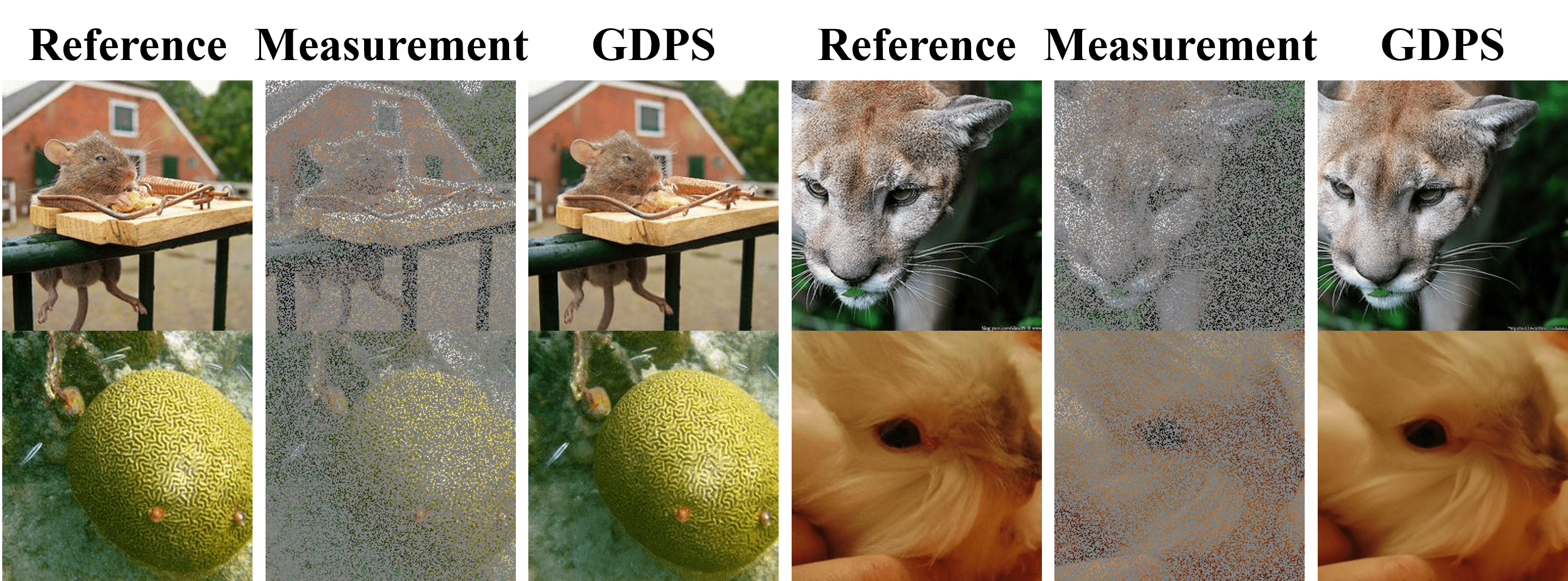}
         \caption{Random Inpainting(70\%)}
     \end{subfigure}
     \begin{subfigure}[b]{0.6\textwidth}
         \centering
         \includegraphics[width=\textwidth]{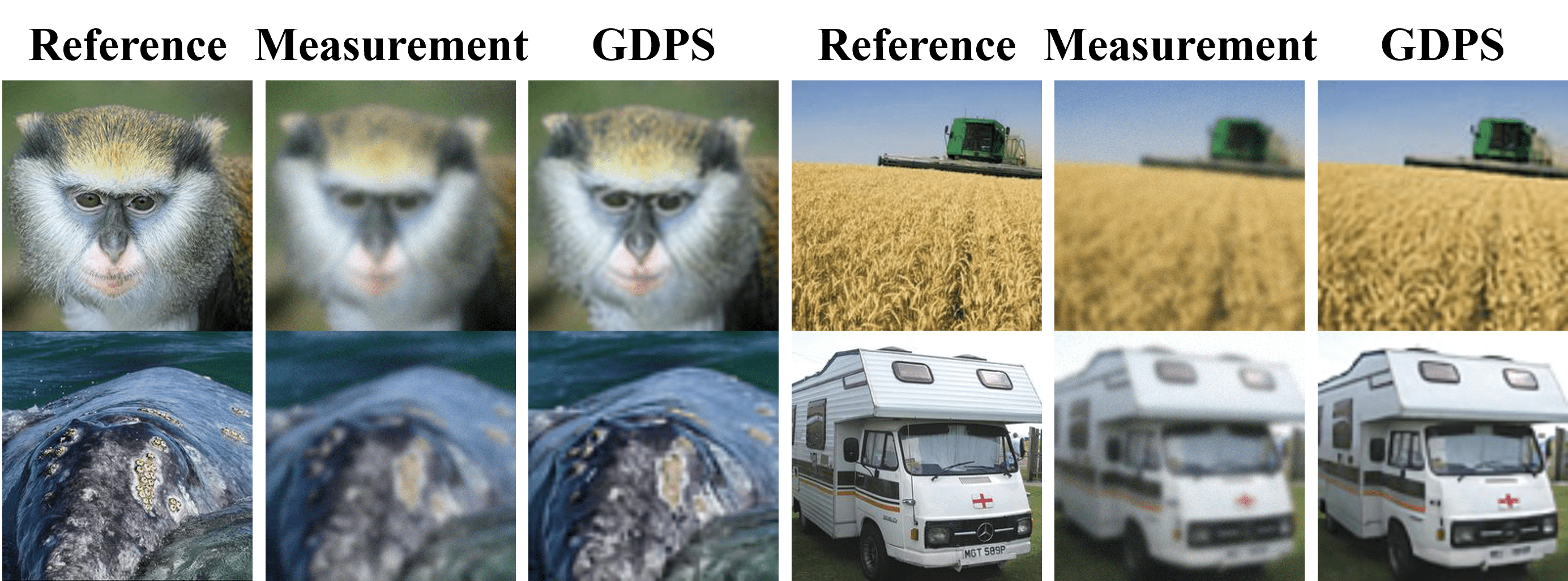}
         \caption{Gaussian Deblurring}
     \end{subfigure}
     \begin{subfigure}[b]{0.6\textwidth}
         \centering
         \includegraphics[width=\textwidth]{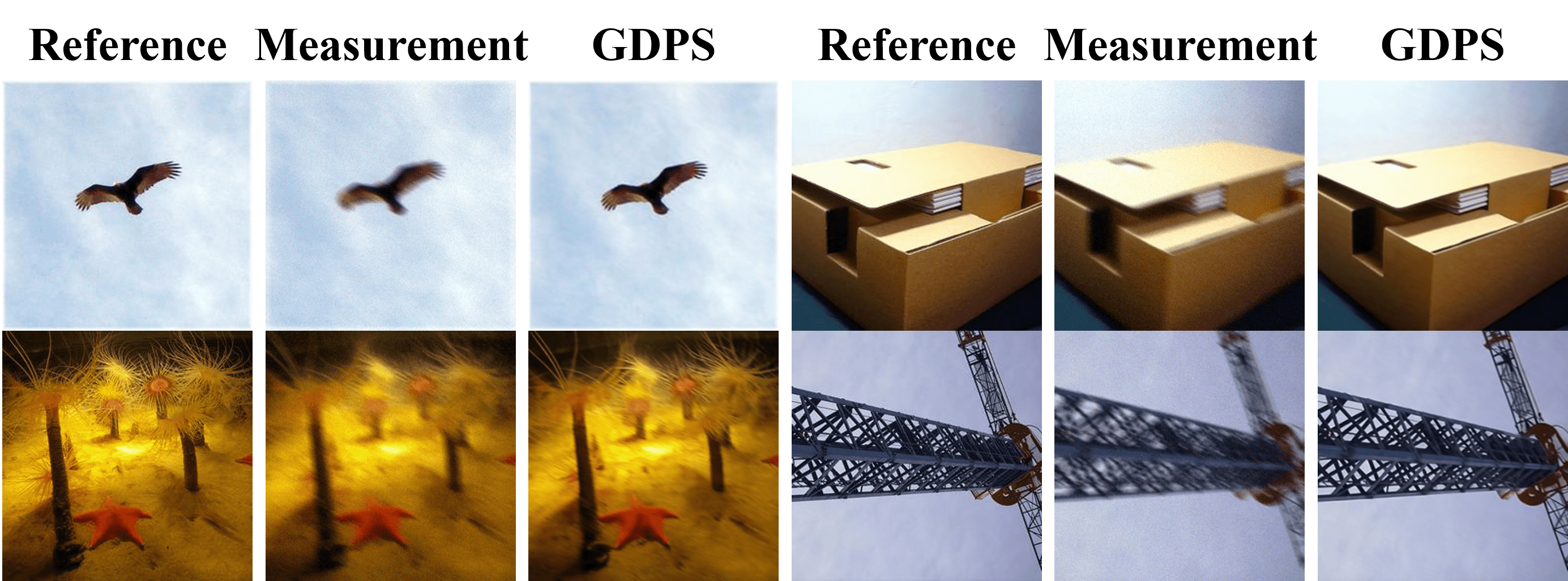}
         \caption{Motion Deblurring}
     \end{subfigure}
        \caption{More results from the experiments in the validation set of \textbf{ImageNet 256x256} dataset. Arranged from top to bottom, the taks are: super resolution, box inpainting, random inpainting, gaussian deblurring and motion deblurring.}
\label{fig-ffhq}
\end{figure*}

\newpage

\begin{figure*}[!htbp]
     \centering
     \begin{subfigure}[b]{0.6\textwidth}
         \centering
         \includegraphics[width=\textwidth]{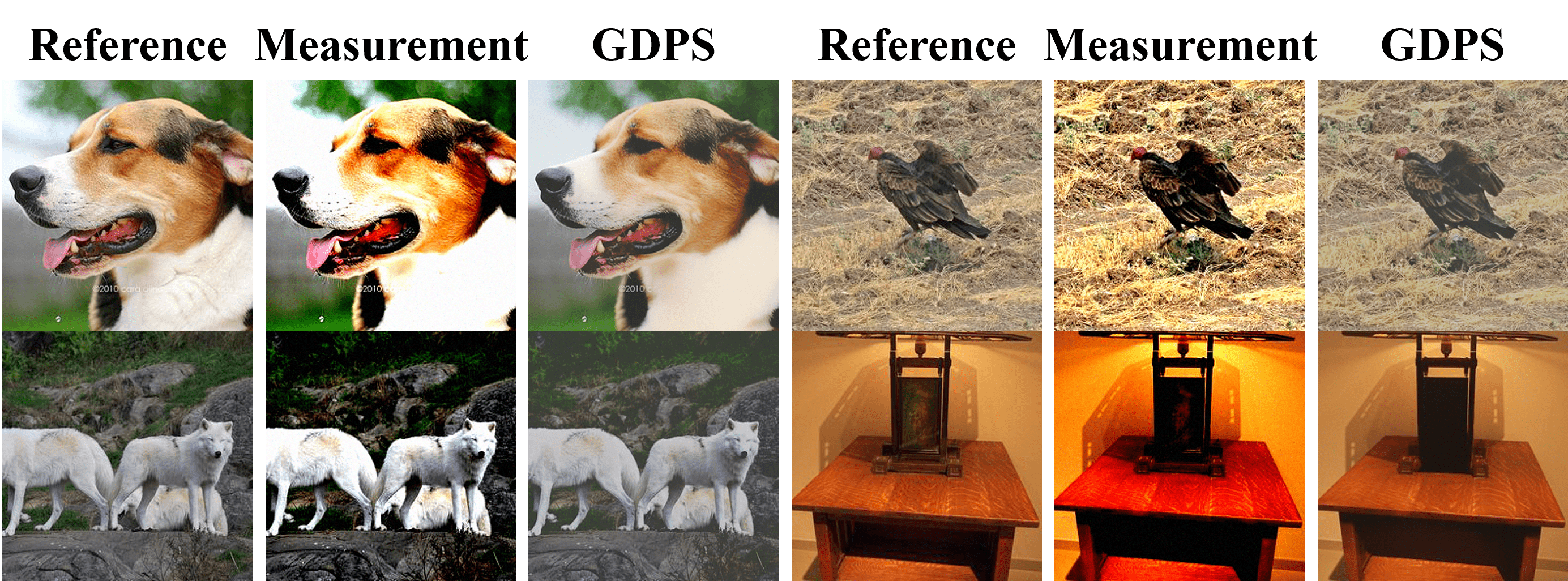}
         \caption{High Dynamic Range}
     \end{subfigure}
     \begin{subfigure}[b]{0.6\textwidth}
         \centering
         \includegraphics[width=\textwidth]{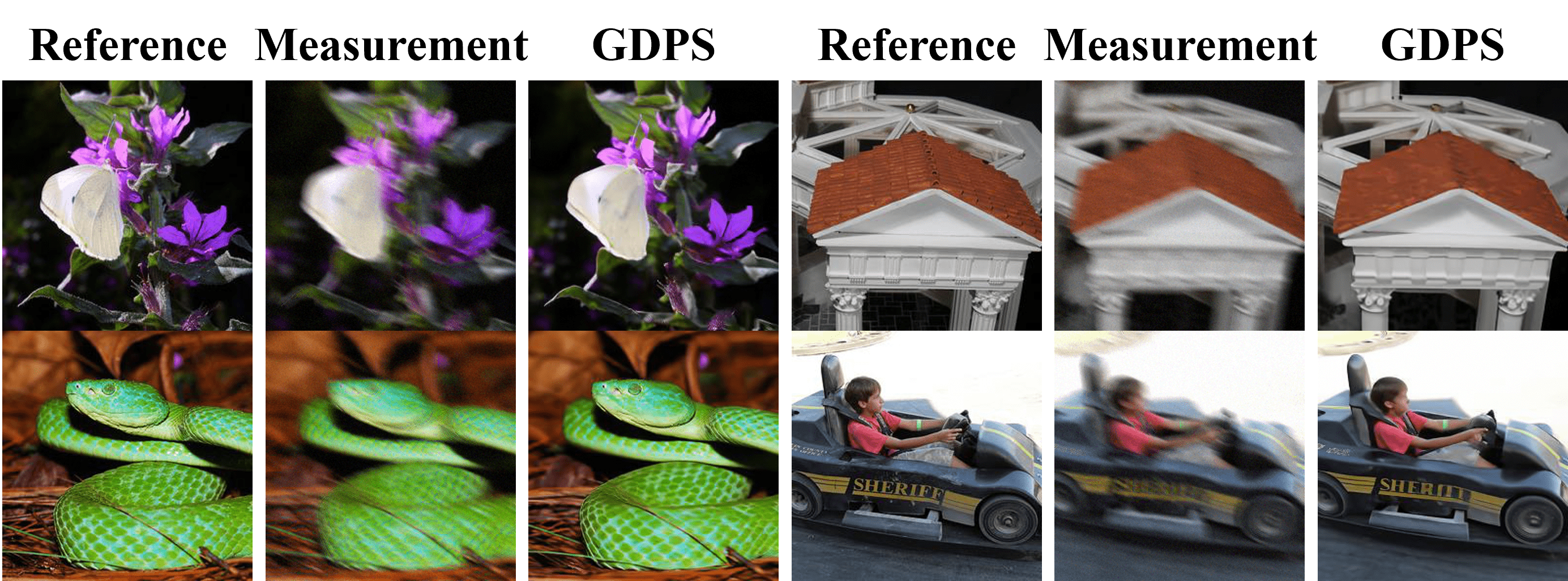}
         \caption{Nonlinear Deblurring}
     \end{subfigure}
     \begin{subfigure}[b]{0.6\textwidth}
         \centering
         \includegraphics[width=\textwidth]{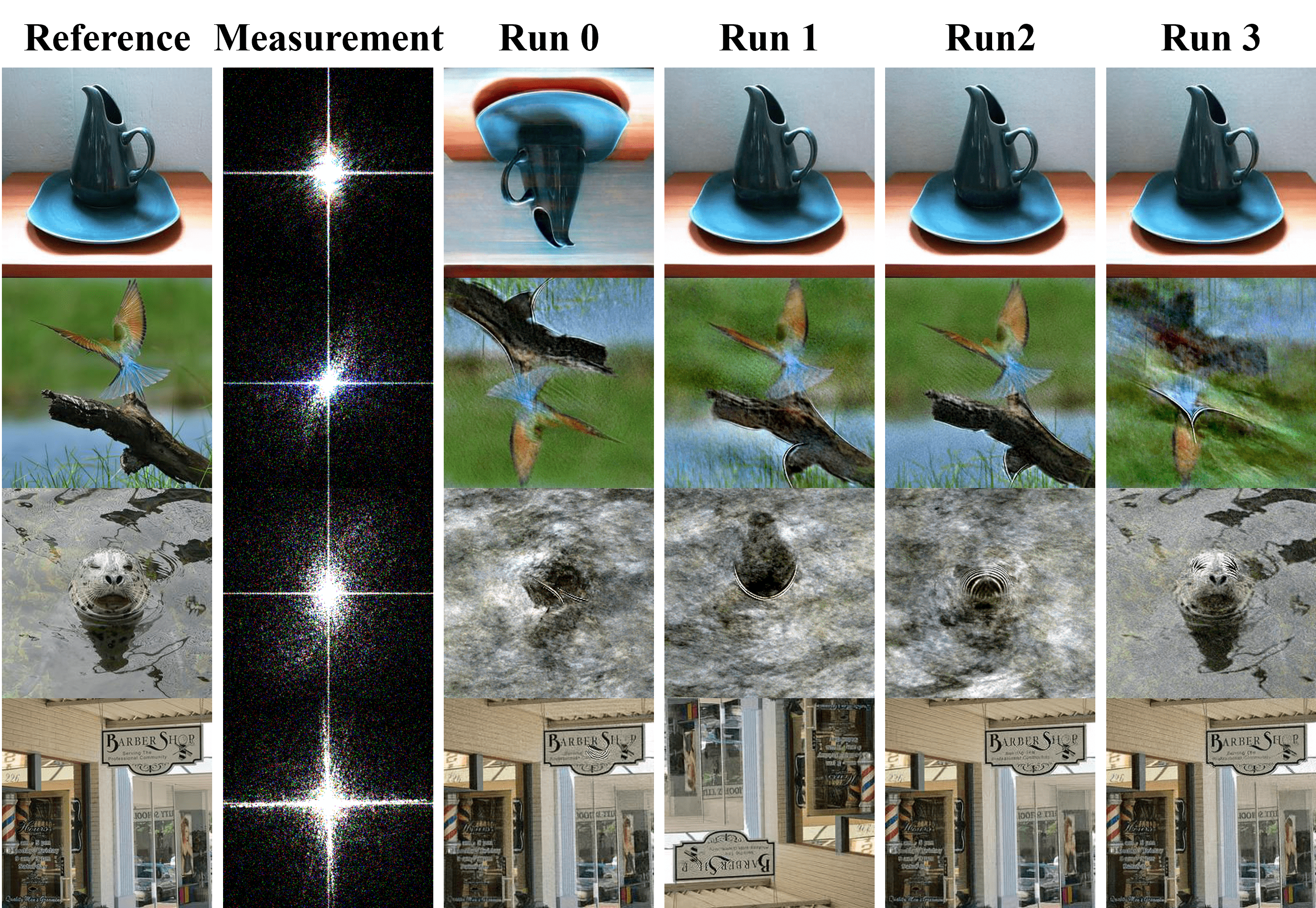}
         \caption{Phase Retrieval}
     \end{subfigure}
        \caption{More results from the experiments in the validation set of \textbf{ImageNet 256x256} dataset. Arranged from top to bottom, the taks are: high dynamic range, nonlinear deblurring and phase retrieval.}
\label{fig-ffhq}
\end{figure*}

\newpage

\begin{figure*}[!htbp]
     \centering
     \begin{subfigure}[b]{0.6\textwidth}
         \centering
         \includegraphics[width=\textwidth]{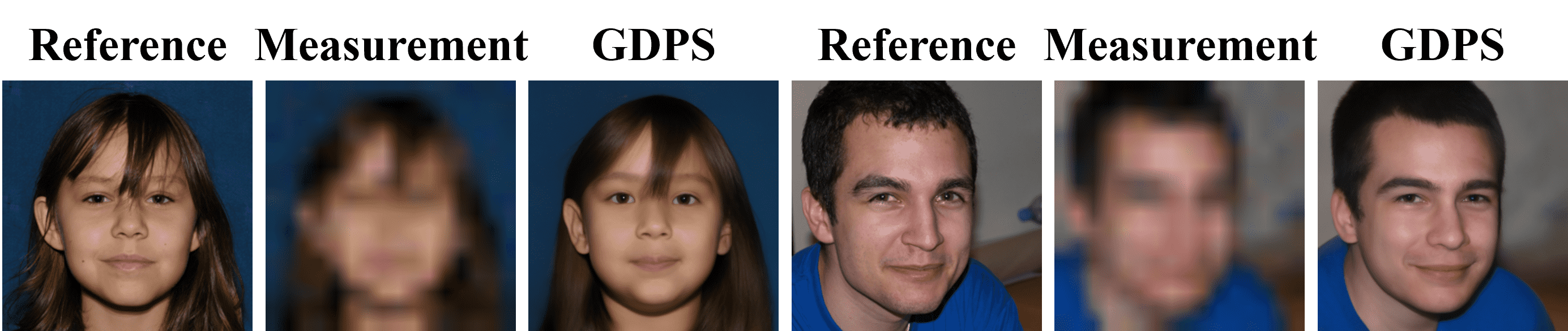}
         \caption{Super Resolution(16x)}
     \end{subfigure}
     \begin{subfigure}[b]{0.6\textwidth}
         \centering
         \includegraphics[width=\textwidth]{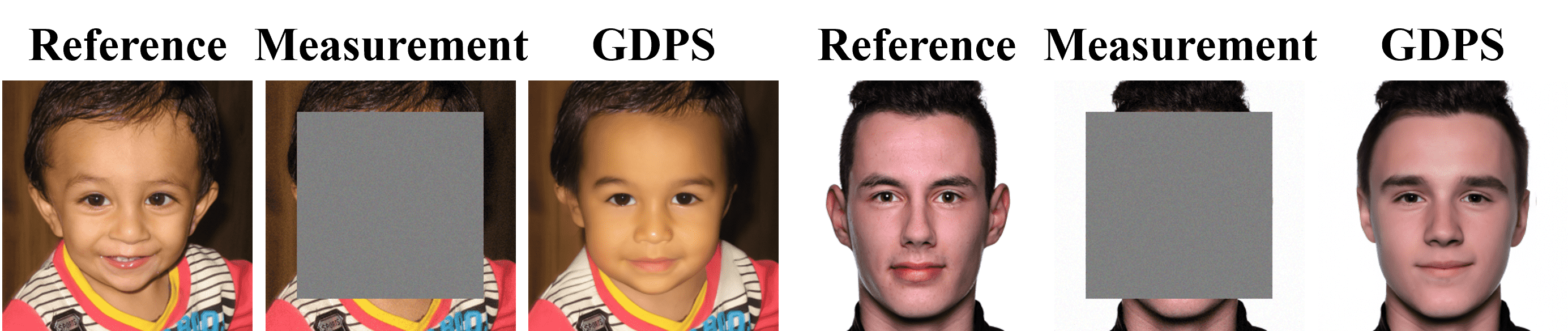}
         \caption{Box Inpainting(192x192)}
     \end{subfigure}
     \begin{subfigure}[b]{0.6\textwidth}
         \centering
         \includegraphics[width=\textwidth]{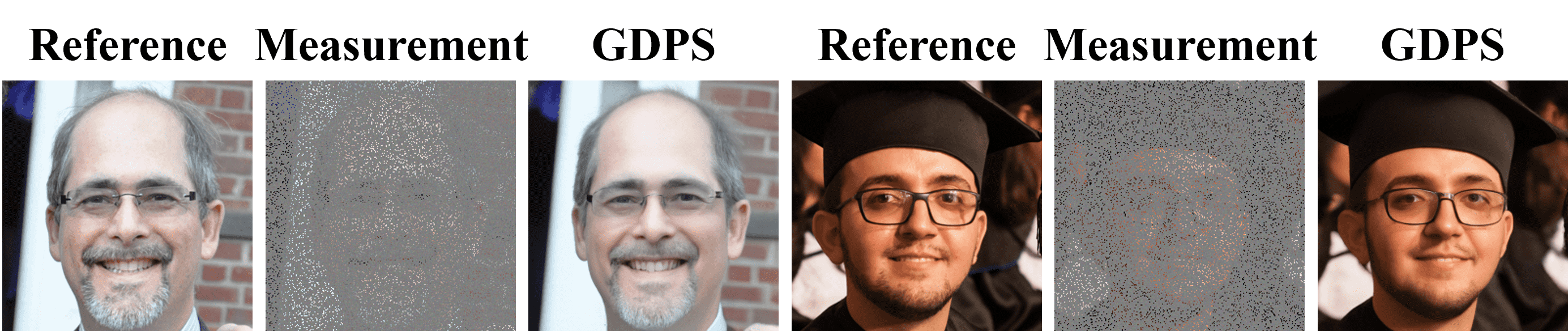}
         \caption{Random Inpainting(90\%)}
     \end{subfigure}
     \begin{subfigure}[b]{0.6\textwidth}
         \centering
         \includegraphics[width=\textwidth]{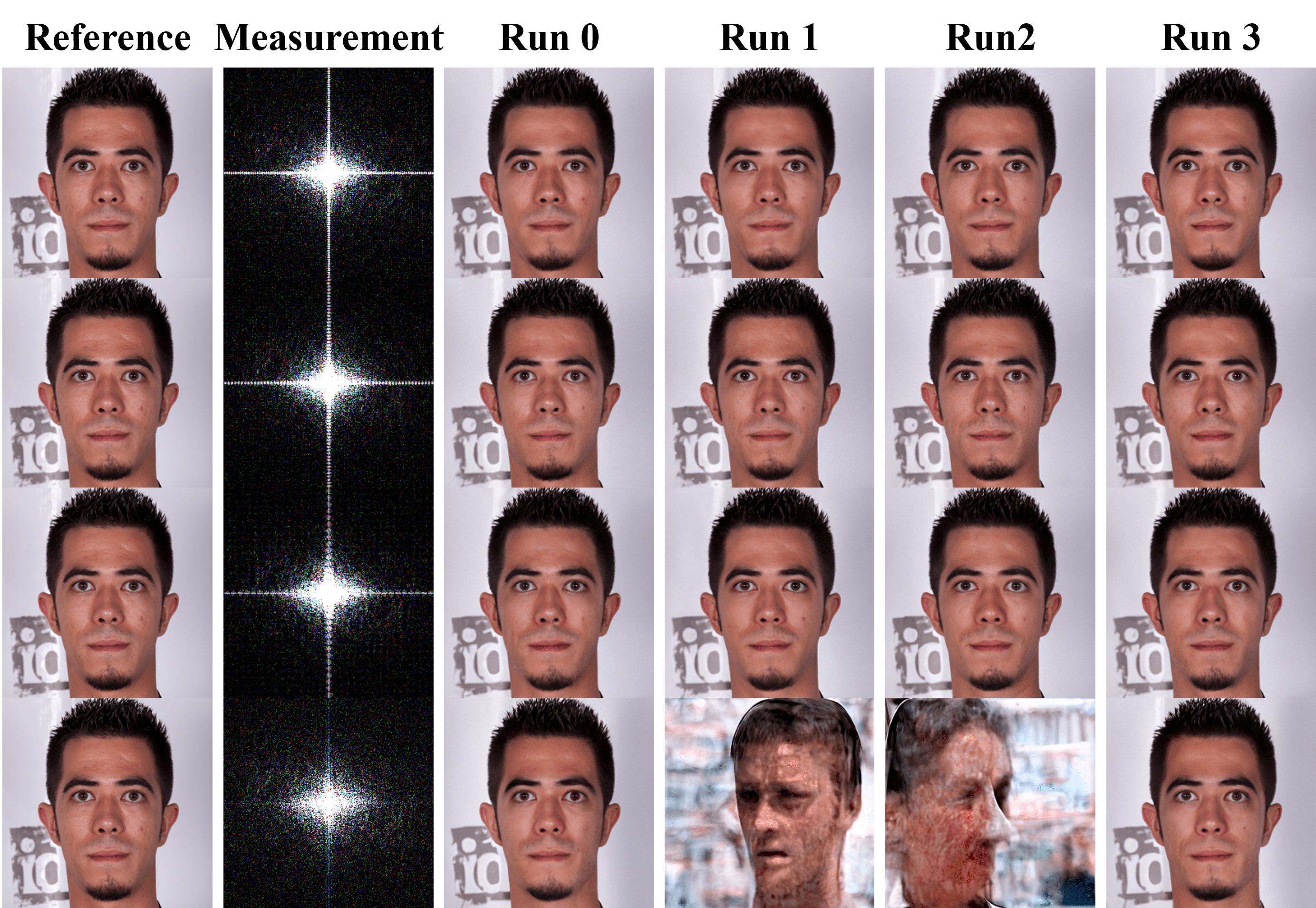}
         \caption{Phase Retrieval}
     \end{subfigure}
        \caption{More results from the experiments in the validation set of \textbf{FFHQ 256x256} dataset. Arranged from top to bottom, the taks are: super resolution(16x), box inpainting(192x192), random inpainting(90\%) and phase retrieval. In the phase retrieval task, the oversampling factors, listed from top to bottom are: 1.5, 1.0, 0.5, and 0.0.}
\label{fig-ffhq}
\end{figure*}

\end{document}